\pdfoutput=1

\documentclass[11pt]{article}

\usepackage[preprint]{acl}

\usepackage{times}
\usepackage{latexsym}

\usepackage[T1]{fontenc}
\usepackage[utf8]{inputenc}

\usepackage{microtype}
\usepackage{inconsolata}

\usepackage{graphicx}

\usepackage{breqn}
\usepackage{amsmath}
\usepackage{amssymb}
\usepackage{mathtools}
\usepackage{amsthm}
\usepackage{enumitem} 
\usepackage{colortbl} 
\usepackage{xcolor}   

\usepackage{microtype}
\usepackage{graphicx}
\usepackage{subcaption}
\usepackage{booktabs} 
\usepackage{caption}   
\usepackage{subcaption} 
\usepackage{float}      
\usepackage{bbding}
\usepackage{algorithm} 
\usepackage{algorithmic}

\usepackage{hyperref}
\definecolor{Gray}{gray}{0.7} 
\usepackage[capitalize,noabbrev]{cleveref}

\usepackage{listings}
\usepackage{xcolor}
\definecolor{textblue}{rgb}{.2,.2,.7}
\definecolor{textred}{rgb}{0.54,0,0}
\definecolor{textgreen}{rgb}{0,0.43,0}
\usepackage{listings}
\title{Speculative Decoding Scaling Laws (SDSL): \\Throughput Optimization Made Simple}

\author{
Amirhossein Bozorgkhoo\thanks{Independent Researcher}
\and
Igor Molybog\thanks{University of Hawai’i at Manoa, Honolulu, HI, USA.}
}

\begin{document}
\maketitle
\begin{abstract}
Speculative decoding is a technique that uses multiple language models to accelerate inference. Previous works have used an experimental approach to optimize the throughput of the inference pipeline, which involves LLM training and can be costly.
This study of speculative decoding proposes a theory that analytically connects the key hyperparameters of pre-trained LLMs to the throughput efficiency of a downstream SD-based inference system. The theory allows the prediction of throughput-optimal hyperparameters for the components of an inference system before their pre-training.
\end{abstract}

\section{Introduction}
\label{introduction}

Speculative decoding is an effective technique to accelerate Large Language Model (LLM) inference~\citep{chen2023accelerating}. It uses a smaller draft model to sequentially generate multiple candidate tokens, which the target model verifies in parallel. Speculative decoding increases throughput while maintaining accuracy. However, the success of this approach is \textit{highly dependent on the choice of the draft model}—an ill-suited draft model can introduce latency bottlenecks, reducing or negating the speedup benefits of speculative decoding.
Current methods for selecting an appropriate draft model rely on empirical search and benchmarking across multiple architectures, requiring extensive computational resources and research efforts \citep{chen2023accelerating, yan2024decodingspeculativedecoding}. An overview of the most related literature is provided in Appendix \ref{sec:related}.

We propose an \textit{analytical framework (scaling law)} for deriving the optimal draft model size ahead of training, assuming it will be trained on a similar dataset to the one used for the target model. This framework is designed to systematically balance draft model inference latency and its accuracy in language modeling tasks. 

The main contributions of the paper are:

\begin{itemize}
    \item We establish a simple analytical relationship of the form
    $$\alpha = a x + b y + c$$
    between draft model perplexity $x$, target model perplexity $y$,  and their "alignment" $\alpha$ (the expected token acceptance). 

    \item Assuming that both draft and target models are pre-trained from scratch, we derive a numerical relationship of the form
    $$N_{\text{opt}} = M_0 + \mu M$$
    between the size of the target model $M$ and the optimal draft model $N_{\text{opt}}$. We find that the draft model should be approximately two orders of magnitude ($200$x) smaller than the target model, and this relationship remains robust across different model families.

    \item Assuming both models are trained on a comparable scale (on the order of a trillion tokens), the impact of dataset size on throughput remains mild.
\end{itemize}

We establish a broader Speculative Decoding Scaling Law (SDSL) framework. This framework allows practitioners to adapt the pre-training scaling laws—those connecting perplexity to model size or training dataset properties—for a principled selection of draft models without the need for an additional exhaustive empirical search.
The models trained for establishing the pre-training scaling laws can be reused to study the SDSL coefficients $a$ and $b$ for a specific model family. Following our protocol, one can obtain particular values of $M_0$ and $\mu$ without conducting additional experiments and decide on the draft model architecture, given a specific target model size they already have in mind.
Our framework is developed for throughput measured in token/FLOP, but we validate this further with measuring token/sec and compare with token/FLOP results. 

\section{Background}
\label{background}

\subsection{Speculative Decoding} 
\label{sec:SD}

Speculative Decoding utilizes a dual-model architecture to enhance the throughput of token generation in natural language generation tasks. Let \( M_p \) denote the \textbf{target} language model we would like to use for inference. It outputs the probability distribution over the vocabulary of tokens \( p(x_t | x_{<t}) \) for a given prefix \( x_{<t} \). In contrast, the \textbf{draft} language model \( M_q \) is a smaller and more compute-efficient language model that could be used for the same task of probability estimation represented by \( q(x_t | x_{<t}) \) \citep{leviathan2023fastinferencetransformersspeculative}.

The core strategy of Speculative Decoding involves several key steps:
\begin{enumerate}
    \item  Use the efficient model \( M_q \) to generate \( \gamma \in \mathbb{Z}^+ \) completions.
    \item Evaluate all generated completions in parallel using the target model \( M_p \). Accept or reject tokens based on a threshold rule.
    \item Sample an additional token from an adjusted distribution if the first one is rejected, or add an additional token if all completions are accepted.
\end{enumerate}

For the cost of only running \( M_p \) once, this approach can potentially generate up to \( \gamma + 1 \) new tokens—depending on how well \( M_q \) approximates \( M_p \). It also ensures that the sampled tokens are drawn exactly according to the target distribution $p.$ The lookahead length \( \gamma \) represents the maximum number of tokens generated by the draft model.

\paragraph{Expected acceptance \textbf{$\alpha$}}

In speculative decoding, a token \( x_t \) generated from the distribution \( q(x_t | x_{<t}) \) can be accepted or rejected according to a threshold rule on \( p(x_t | x_{<t}) \). The acceptance rate \( \beta_{x_{<t}} \) is defined as the probability of accepting a token \( x_t \) sampled from the distribution \( q(x_t | x_{<t}) \), given a prefix \( x_{<t} \). 
To quantify effectiveness of speculative decoding, the parameter 
\[ \alpha = \mathbb E_{x_{<t}}(\beta_{x_{<t}}) \] 
is introduced, representing the expected acceptance rate across prefixes. The value of \( \alpha \) measures how well the draft distribution aligns with the target distribution \citep{leviathan2023fastinferencetransformersspeculative}. The practical method we use for estimating $\alpha$ is provided in Appendix \ref{sec:alpha_CI}.

\paragraph{Throughput Enhancement in Speculative Decoding
}\citet{leviathan2023fastinferencetransformersspeculative} analyze how their speculative decoding algorithm enhances inference efficiency by reducing wall-clock time. They demonstrate that, under the assumption of independent and identically distributed (i.i.d.) inputs, their method decreases the number of calls to the target model, \(M_p\), by a factor of \(\frac{1 - \alpha^{\gamma + 1}}{1 - \alpha}\). This reduction is contingent on having sufficient computational resources to support increased concurrency, allowing for \(\gamma + 1\) concurrent evaluations of \(M_p\) without extending wall-clock time.

To quantify wall-clock time improvement, they introduce a cost coefficient \(c\), which represents the ratio of the time taken for a single run of the approximation model \(M_q\) relative to that of \(M_p\). In their experiments, \(c\) is consistently less than 0.05, indicating that \(M_q\) is significantly smaller and faster than \(M_p\). The authors derive a theorem stating that the expected improvement factor in total wall-clock time using their algorithm is given by
\begin{equation}\label{eq:improv_factor}
   \frac{\text{Spec dec Throughput}}{M_p \text{Throughput}} = \frac{1 - \alpha^{\gamma + 1}}{(1 - \alpha)(\gamma c + 1)}.
\end{equation} 
Furthermore, they establish a corollary indicating that if \(\alpha > c\), there exists a value of \(\gamma\) that leads to an improvement in wall-clock time, with a minimum improvement factor of at least \(\frac{1 + \alpha}{1 + c}\). This analysis underscores the potential for significant throughput improvements in practical applications of their speculative decoding approach, particularly when computational resources are adequately provisioned. 

However, the connection between $c$ and $\alpha$ observed for modern language models remains unexplored, which is the gap our work is aiming to fill.

\subsection{Scaling Laws for Pre-training}
\label{sec:Scaling Laws}

Scaling laws were previously developed to inform decision-making during the design stage of large-scale training experiments.
The scaling laws for pre-training of large language models analytically relate the key metrics of the trained model performance, such as cross-entropy loss $L$, to the key hyperparameters of the training routine, such as the number of model parameters $ N$ and the number of training tokens  $ D.$ A common way to capture the dependency is in the form suggested by \citet{hoffmann2022trainingcomputeoptimallargelanguage}
 \begin{equation}
	\label{eq:Perplexity}
	L(N,D) = \ln x(N, D) = E + \frac{A}{N^{\nu}} + \frac{B}{D^{\delta}},
\end{equation}
The constants $ E $, $ A $, $ B $, $ \nu $, and $ \delta $ are parameters that capture the irreducible loss and the diminishing returns on loss reduction as model size and training data increase. Specifically, $ A/N^{\nu} $ models the effect of increasing model size on loss, while $ B/D^{\delta} $ accounts for the impact of additional training tokens.

Similarly, the presented work aims to develop a scaling law of an inference system to inform decision-making during the design stage of an AI service.

\section{Methodology}
\label{sec:methodology}

\subsection{FLOPs of a Speculative Decoding iteration}

We first compute throughput in terms of tokens per inference FLOP, which allows us to abstract away from any specific hardware configuration. This formulation is motivated by the strong correlation between inference wall-clock time and the computational workload of the inference process. We then validate this approach by measuring wall-clock latency across draft–target pairs. We find this methodology to be more objective, as it naturally adapts to rapid advances in hardware performance.

As shown by~\citet{kaplan2020scalinglawsneurallanguage}, the total number of floating point operations (FLOPs) for a transformer model during a forward pass with a small context length can be efficiently approximated by \(2N\), where \(N\) is the number of parameters (size) of the model. Here and further in the paper, the size of the target model is denoted as \(M\), while \(N\) is reserved for the size of the draft model. During each iteration, the draft model generates \(\gamma\) speculative guesses, requiring \(2N\gamma\) FLOPs for a forward pass through the draft model. Additionally, for each guess generated, the target model evaluates these guesses in parallel, incurring another \(2M\) FLOPs. Therefore, the total computational cost of a single speculative decoding iteration is $2(M+ \gamma \cdot N).$


\subsection{Modeling Throughput}
\label{sec:throughput}
\citet{yan2025decodingspeculativedecoding} presented a formula for the throughput of a speculative decoding system expressed in tokens per second. We adapt their approach to express throughput in tokens per FLOP.
We accept $c=\frac{N}{M}$ and use \eqref{eq:improv_factor} to establish that the throughput of speculative decoding is proportional to
\begin{equation}
   \label{eq:Throughput_orginal}
\mathcal{T} = \frac{1 - \alpha^{\gamma+1}}{2(M + \gamma\cdot N)(1 - \alpha)},
\end{equation}
with an architecture-defined proportionality coefficient.

The value of lookahead length $\gamma$ can be set arbitrarily, complicating the analysis of optimal throughput. To abstract this hyperparameter out, we assume the value of $\gamma$ to be optimally chosen. In Appendix \ref{sec:gamma_derivation}, we show that the throughput under optimal $\gamma$ is captured through 
{
\begin{equation}
   \label{eq:Throughput}
   \mathcal{T} = \frac{-\log(\alpha)}{2 N (-1 + \alpha) W\left(-\frac{\alpha^{M/N-1}}{e}\right)} 
\end{equation}}
where \( W \) denotes the Lambert W function. The dependency of $\alpha$ on the other hyperparameters is the focus of the experimental part of our work.

\begin{table*}[ht]
    \centering
     \resizebox{\textwidth}{!}{ 
      \begin{tabular}{|c|c|c|c|c|c|c|c|c|c|} 
         \hline
         \quad & \textbf{Draft Model}  \quad  & OPT-125M & OPT-350M & OPT-1.3B & OPT-2.7B  \\ 
         \hline
         \textbf{Target Model} & Perplexity & 29.79318619 &  25.32799339   &  19.36876869 &  17.36415863  \\ 
         \hline
          OPT-13B& 15.58453751 & ${0.5959}_{\pm 0.0018}$ & ${0.6281}_{\pm 0.0017}$ & ${0.6694}_{\pm 0.0016}$ & ${0.6793}_{\pm 0.0016}$ \\ 
         \hline
          OPT-30B& 15.30226898 & ${0.6779}_{\pm 0.0017}$ & ${0.6984}_{\pm 0.0016}$ & ${0.7325}_{\pm 0.0015}$  & ${0.7447}_{\pm 0.0015}$ \\ 
         \hline
          OPT-66B& 14.58085632 & ${0.7118}_{\pm 0.0016}$ & ${0.7342}_{\pm 0.0015}$ & ${0.7659}_{\pm 0.0014}$ & ${0.7847}_{\pm 0.0014}$ \\ 
         \hline
          Qwen1.5-14B& 11.7312355 & ${0.599}_{\pm 0.0013}$ & ${0.6178}_{\pm 0.0012}$ & ${ 0.6569}_{\pm 0.0012}$ & ${0.6708}_{\pm 0.0012}$\\ 
         \hline
         Qwen1.5-32B& 10.41608143 & ${0.5842}_{\pm 0.0013}$ & ${0.6086}_{\pm 0.0013}$ & ${ 0.6458}_{\pm 0.0012}$ & ${0.6617}_{\pm 0.0012}$\\ 
         \hline
         Qwen1.5-72B& 10.33209324 & ${0.5929}_{\pm 0.0013}$ & ${0.6177}_{\pm 0.0013}$ & ${ 0.6511}_{\pm 0.0012}$ & ${0.6658}_{\pm 0.0012}$\\ 
         \hline
         Qwen1.5-110B& 10.07259178 & ${0.5206}_{\pm 0.0013}$ & ${0.5461}_{\pm 0.0013}$ & ${ 0.5816}_{\pm 0.0013}$ & ${0.5981}_{\pm 0.0013}$\\ 
         \hline
         Qwen2.5-14B& 10.29567051 & ${0.5998}_{\pm 0.0021}$ & ${0.6184}_{\pm 0.002}$ & ${ 0.653}_{\pm 0.0019}$ & ${0.6651}_{\pm 0.0019}$\\ 
         \hline
         Qwen2.5-32B& 9.549574852 & ${0.5755}_{\pm 0.0019}$ & ${0.5954}_{\pm 0.0018}$ & ${ 0.632}_{\pm 0.0017}$ & ${0.6495}_{\pm 0.0017}$\\ 
         \hline
         Qwen2.5-72B& 10.11350441 & ${0.5702}_{\pm 0.0019}$ & ${0.5917}_{\pm 0.0019}$ & ${ 0.6245}_{\pm 0.0018}$ & ${0.6443}_{\pm 0.0018}$\\ 
         \hline
         LLaMa3-70B& 11.00730228 & ${0.6027}_{\pm 0.0014}$ & ${0.6282}_{\pm 0.0013}$ & ${ 0.6605}_{\pm 0.0013}$ & ${0.6744}_{\pm 0.0013}$\\ 
         \hline
         LLaMa3.1-70B& 11.07195759 & ${0.6057}_{\pm 0.0018}$ & ${0.6246}_{\pm 0.0018}$ & ${ 0.6647}_{\pm 0.0017}$ & ${0.6738}_{\pm 0.0017}$\\ 
         \hline
         Seed-OSS-36B& 9.997225761 & ${0.5347}_{\pm 0.0022}$ & ${0.557}_{\pm 0.0022}$ & ${ 0.5861}_{\pm 0.0021}$ & ${0.6033}_{\pm 0.0021}$\\ 
         \hline
      \end{tabular}
     }
     \caption{Estimated $\alpha$ values along with their associated 95\% confidence intervals  (associated with stochasticity in evaluation and not training) and measured perplexity for selected pairs of all target models  and OPT draft models.}
    \label{Table:Alpha_perplexity:OPT_drafts}
  
\end{table*}

\begin{table*}[ht]

\centering
  \scriptsize
  \resizebox{\textwidth}{!}{ 
      \begin{tabular}{|c|c|c|c|c|c|c|c|c|} 
         \hline
         & \textbf{Draft Model} & Qwen2.5-0.5B & Qwen2.5-1.5B & Qwen2.5-3B & Qwen1.5-0.5B & Qwen1.5-1.8B & Qwen1.5-4B  \\ 
         \hline
          \textbf{Target Model} & Perplexity
          & 17.91464424 & 13.69491291  & 12.5883894 & 18.37218094 & 14.10007191 & 13.90737915  \\ 
         \hline			
	       OPT-13B & 15.58453751 & ${0.6516}_{\pm 0.0017}$ & ${0.6678}_{\pm 0.0017}$  & ${0.6725}_{\pm 0.0016}$ & ${0.649}_{\pm 0.0017}$ & ${0.6672}_{\pm 0.0017}$ & ${0.6716}_{\pm 0.0017}$\\
          \hline
          OPT-30B & 15.30226898 & ${0.7282}_{\pm 0.0016}$ & ${0.7414}_{\pm 0.0015}$ & ${0.749}_{\pm 0.0015}$ & ${0.7241}_{\pm 0.0016}$ & ${0.7372}_{\pm 0.0015}$ &  ${0.7483}_{\pm 0.0015}$\\
         \hline
         OPT-66B & 14.58085632 &  ${0.7678}_{\pm 0.0014}$ & ${0.7881}_{\pm 0.0013}$ & ${0.7882}_{\pm 0.0013}$ & ${0.7634}_{\pm 0.0013}$ & ${0.7808}_{\pm 0.0014}$ &  ${0.7875}_{\pm 0.0014}$\\ 
         \hline
          Qwen1.5-14B & 11.7312355  & ${0.6898}_{\pm 0.0011}$ & ${0.7244}_{\pm 0.0011}$ & ${0.7323}_{\pm 0.0011}$ & ${0.6838}_{\pm 0.0011}$ & ${0.7131}_{\pm 0.0011}$ &  ${0.7343}_{\pm 0.0011}$\\
          \hline
          Qwen1.5-32B & 10.41608143 & ${0.6698}_{\pm 0.0012}$ & ${0.7069}_{\pm 0.0011}$ & ${0.7152}_{\pm 0.0011}$ & ${0.6585}_{\pm 0.0012}$ & ${0.6873}_{\pm 0.0012}$ &  ${0.7084}_{\pm 0.0011}$\\
         \hline
          Qwen1.5-72B & 10.33209324 & ${0.6685}_{\pm 0.0012}$ & ${0.7065}_{\pm 0.0011}$ & ${0.7156}_{\pm 0.0011}$ & ${0.6607}_{\pm 0.0012}$ & ${0.6848}_{\pm 0.0012}$ &  ${0.7079}_{\pm 0.0011}$\\
         \hline
          Qwen1.5-110B & 10.07259178  & ${0.6064}_{\pm 0.0012}$ & ${0.6399}_{\pm 0.0012}$ & ${0.6503}_{\pm 0.0012}$ & ${0.5907}_{\pm 0.0013}$ & ${0.6215}_{\pm 0.0012}$ &  ${0.6401}_{\pm 0.0012}$\\
         \hline
         Qwen2.5-14B & 10.29567051 & ${0.6783}_{\pm 0.0013}$ & ${0.7109}_{\pm 0.0012}$ & ${0.7202}_{\pm 0.0012}$ & ${0.6742}_{\pm 0.0019}$ & ${0.6878}_{\pm 0.0018}$ &  ${0.7089}_{\pm 0.0017}$\\ 
         \hline
          Qwen2.5-32B & 10.11350441  & ${0.6698}_{\pm 0.0012}$ & ${0.7111}_{\pm 0.0011}$ & ${0.722}_{\pm 0.0011}$ & ${0.6651}_{\pm 0.0017}$ & ${0.6863}_{\pm 0.0016}$ &  ${0.7062}_{\pm 0.0016}$\\
          \hline
          Qwen2.5-72B & 9.549574852 & ${0.6663}_{\pm 0.0012}$ & ${0.7079}_{\pm 0.0011}$ & ${0.7139}_{\pm 0.0011}$ & ${0.6525}_{\pm 0.0018}$ & ${0.6757}_{\pm 0.0017}$ &  ${0.7006}_{\pm 0.0016}$\\
         \hline
          LLaMa3-70B & 11.00730228 & ${0.6959}_{\pm 0.0012}$ & ${0.7367}_{\pm 0.0011}$ & ${0.7408}_{\pm 0.0011}$ & ${0.6815}_{\pm 0.0017}$ & ${0.705}_{\pm 0.0017}$ &  ${0.7283}_{\pm 0.0016}$\\
         \hline
         LLaMa3.1-70B & 11.07195759 & ${0.6699}_{\pm 0.0017}$ & ${0.7035}_{\pm 0.0016}$ & ${0.7149}_{\pm 0.0016}$ & ${0.6584}_{\pm 0.0017}$ & ${0.6816}_{\pm 0.0017}$ &  ${0.6998}_{\pm 0.0016}$\\
         \hline
         Seed-OSS-36B & 9.997225761 & ${0.5998}_{\pm 0.0021}$ & ${0.6313}_{\pm 0.002}$ & ${0.6422}_{\pm 0.002}$ & ${0.5855}_{\pm 0.0021}$ & ${0.6079}_{\pm 0.0021}$ &  ${0.6264}_{\pm 0.002}$\\
         \hline
     \end{tabular}
    }
        \caption{Estimated $\alpha$ values along with their associated 95\% confidence intervals  (associated with stochasticity in evaluation and not training) and measured perplexity for selected pairs of all target models and Qwen draft models.}
     \label{Table:Alpha_perplexity:Qwen_drafts}

\end{table*}

\section{Experiments}
\label{sec:exp}
 
We aim to derive an analytical expression that relates the throughput of a speculative decoding system to the training hyperparameters of the component models, primarily the sizes of the draft ($N$) and target ($M$) models, but also the amount of data ($D$) they are trained on. To achieve this, we study the dependency of $\alpha$ on the perplexity characteristics of the target and draft models. After that, we use the pre-training scaling laws relating the perplexity metrics to the model hyperparameters and combine them with  \eqref{eq:Throughput}. 

We evaluate our framework on a diverse set of LLMs, spanning multiple model families, including LLaMA 3~\cite{2024arXiv240721783G}, LLaMA 3.1~\cite{meta_llama3_1}, OPT~\cite{opt}, Qwen 1.5~\cite{qwen}, Qwen 2.5~\cite{qwen2.5}, and an open-source model released by ByteDance Seed~\cite{seed2025seed-oss}.

\subsection{Estimating \(\alpha\) values}
\label{α_ppl}
\paragraph{Setup.} We utilize the Microsoft Deepspeed library~\cite{deepspeed} to implement speculative decoding, following the methodology discussed in Appendix \ref{sec:alpha_CI}.

The target and draft model pairs used in our experiments are indicated in Tables \ref{Table:Alpha_perplexity:OPT_drafts} and \ref{Table:Alpha_perplexity:Qwen_drafts}. We evaluate the perplexity of these models on the HellaSwag~\cite{hellaswag} dataset, which consists of commonsense reasoning tasks requiring models to complete sentences based on provided contexts.

We use an open-ended generation strategy, enabling models to produce full responses rather than single-letter answers, better reflecting real-world usage and commonsense reasoning capabilities.

\paragraph{Implementation Framework.} 
	

    In our setup, we generated human-readable ground-truth text from the target, then re-tokenized this output using the draft model's tokenizer. This ensured compatibility and allowed reliable cross-family evaluation. This consistent process was applied across all target–draft pairs for Tables \ref{Table:Alpha_perplexity:OPT_drafts} and \ref{Table:Alpha_perplexity:Qwen_drafts}.
	
	\paragraph{Perplexity Calculation.}
    We used Qwen 2.5, Qwen 1.5 and OPT draft models, computing perplexity on HellaSwag with a max sequence length of 2048. Longer documents were truncated accordingly.

	\paragraph{Results.} 
	
    The empirical estimations of the $\alpha$ values and perplexity across various pairs of target and draft models are collected in Tables \ref{Table:Alpha_perplexity:OPT_drafts} and \ref{Table:Alpha_perplexity:Qwen_drafts}. Across all target models, the larger draft models with lower values of perplexity correspond to higher estimated $\alpha$ values. The dependency of $\alpha$ on the target model size is less pronounced, as evident from the experiments conducted for various target models.

\subsection{Regressing \boldmath{$\alpha$} on perplexity of draft and target models}
\label{sec:Power_laws}

\begin{figure}[ht]
    \centering
        \includegraphics[width=\linewidth]{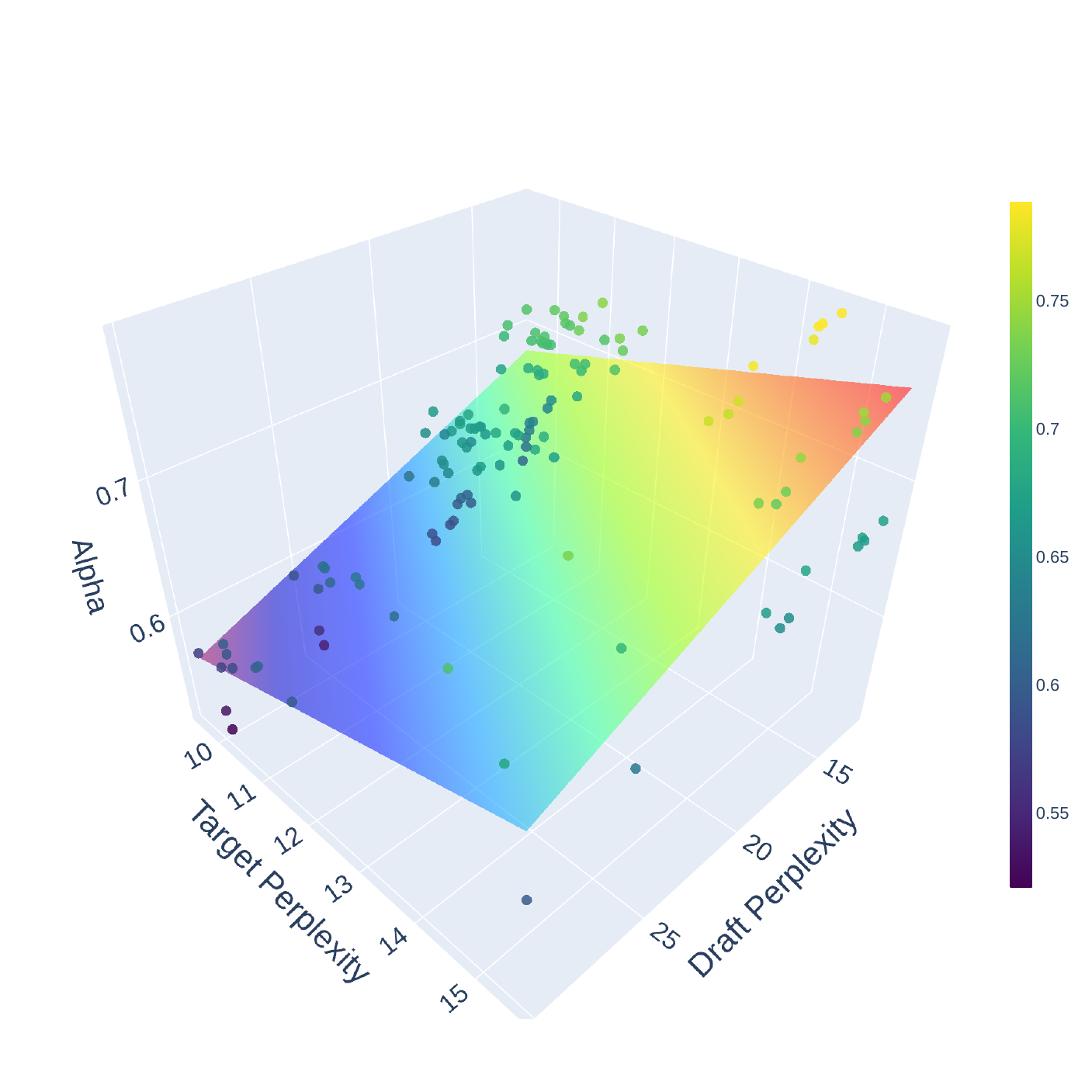}
    \caption{Visualization of the fitted affine plane relating the estimated scaling parameter $\alpha$ to draft model perplexity ($x$) and target model perplexity ($y$). Scatter points represent empirical $(x, y, \alpha)$ observations from Tables~\ref{Table:Alpha_perplexity:OPT_drafts} and~\ref{Table:Alpha_perplexity:Qwen_drafts}, while the surface corresponds to the least-squares fit of Equation~\ref{eq:plane}.}

    \label{fig:fitted_plane}
\end{figure}

 To analyze the relationship between draft model perplexity, target model perplexity and estimated \(\alpha\) values, we fitted an affine plane as a scaling law to the data presented in Tables \ref{Table:Alpha_perplexity:OPT_drafts} and \ref{Table:Alpha_perplexity:Qwen_drafts} : 
 
\begin{equation}
\label{eq:plane}
    \alpha = Ax + By + C
\end{equation}
 Where x is perplexity of the draft model and y is the perplexity of the target model. The estimated values of its parameters, R-squared and Mean Squared Error (MSE) are collected in Table \ref{table:fitted_plane} . The fitted relationship is illustrated in Figure \ref{fig:fitted_plane}.



\paragraph{Discussion.}
The dependence of the scaling parameter $\alpha$ on draft model perplexity and target model perplexity is analyzed separately in Appendix~\ref{sec:regression_on_draft_model} and Appendix~\ref{sec:regression_on_target_model}. Our results show a strong and systematic dependence of $\alpha$ on draft model perplexity, characterized by a consistent monotonic increase in $\alpha$ as draft model perplexity decreases across all target models and fitted functional forms. In contrast, the dependence of $\alpha$ on target model perplexity is comparatively weak and does not exhibit a robust or consistent trend when the draft model is held fixed. 

Despite this asymmetry, we retain both draft and target model perplexities as explanatory variables in the affine scaling law of Equation~\eqref{eq:plane}. While draft model perplexity constitutes the dominant factor governing variations in $\alpha$, incorporating target model perplexity ensures that the scaling law accounts for residual target-dependent effects and remains applicable across a broad range of model families and quality regimes. 
Equation \eqref{eq:plane} captures the dependency of $\alpha$ on the perplexity of the draft model $x$ and perplexity of the target model $y$. Incorporating this into \eqref{eq:Throughput} results in the following formula for throughput:


\begin{dmath}
\label{eq:throughput_on_x}
\mathcal{T}
= \frac{
    -\log(Ax + By + C)
}{
    2N (Ax + By + C - 1)\,
    W\!\left(
        -\frac{(Ax + By + C)^{\frac{M}{N}-1}}{e}
    \right)
}
\end{dmath}

\subsection{Throughput vs size of the draft and target models}
\label{sec:Throughput_Optimal_N}

Working with a specific pre-training dataset and a specific training recipe, one can fit setup-specific scaling law parameters $A$, $B$ and $C,$ since they depend on the properties of the training setup. 

To connect throughput to the training hyperparameters of the component models, we use the scaling law \eqref{eq:Perplexity} that relates the perplexity of a trained model to its size and the size of its training dataset. We first utilize the parameters suggested by \citet{besiroglu2024chinchillascalingreplicationattempt} : 

\begin{equation}\label{eq:perplexity_draft}
x =  e^{1.8172 + \frac{482.01}{N^{-0.3478}} + \frac{2085.43}{D^{-0.3658}}}    
\end{equation}
\begin{equation}
\label{eq:perplexity_target}
y =  e^{1.8172 + \frac{482.01}{M^{-0.3478}} + \frac{2085.43}{D
^{\prime-0.3658}}}    
\end{equation}

where \( D \) and \(D^\prime\) are the number of training tokens used in the draft and target models, respectively. Similarly to $A$, $B$ and $C$, the training setup defines the exact parameter values and could be estimated for a specific training dataset at hand. After substituting these expressions into \eqref{eq:throughput_on_x}, one gets an expression for throughput exclusively through the core parameters of the components of a speculative training system: $M,$ $N,$ and $D$.

{
\begin{dmath}
\label{eq:throughput_N}
\mathcal{T} =  \frac{-\log(Ae^{L(N, D)} + Be^{L(M,D^\prime)} + C)}{2 N (Ae^{L(N, D)} + Be^{L(M,D^\prime)} + C -1)} \cdot \\ \frac{1}{  W\left(\frac{(Ae^{L(N, D)} + Be^{L(M,D^\prime)} +C)^{M/N-1}}{-e}\right)}
\end{dmath}
}
Equation \eqref{eq:throughput_N} captures the dependency of Throughput on the size of the target and draft model along with size of the dataset used to train the target and draft models. 

To examine the relationship between throughput and draft model size \( N \), we numerically analyzed Equation \eqref{eq:throughput_N} using input values from our model families. Through grid search, we estimated the values of \( N \) that yield local throughput maxima, denoted as \( N^\ast \), with results shown in Table~\ref{table:Optimal_draft_model_size_throughput}. We empirically validate this throughput-based prediction by measuring end-to-end inference latency under speculative decoding for an OPT-13B target model, with results reported in Appendix~\ref{sec:validate_optimal_N}.

In Figure~\ref{fig:combined_throughput}, each curve depicts the predicted throughput as a function of the draft model size \(N\) for a fixed target model, with separate panels corresponding to the draft model families. For each target model, the optimal draft size \(N^\ast\) that maximizes throughput is indicated by a star marker placed directly on the curve, while black markers denote the draft model sizes used in our experiments. Across all draft families, throughput initially increases with \(N\), reflecting higher acceptance rates and reduced reliance on the target model.

The observed curvature at the end of the throughput curves is a direct result of the diminishing efficiency of speculative decoding when the draft and target models are similar in size. When \(N \approx M\), the computational savings from using a draft model vanish, since both models require nearly the same amount of FLOPs. Consequently, speculative decoding no longer provides a speedup, and throughput decreases instead of increasing. This results in a negative derivative of throughput with respect to \(N\) near \(N = M\), causing the observed curvature on the right-hand side of the curves.

 \section{Throughput-optimal draft models}
\label{analysis}

This Section is dedicated to the analysis of the draft model size $N^\ast$ that maximizes the predicted value of throughput presented in \eqref{eq:throughput_N}. 
We study in more detail how $N^\ast$ depends on the size of the target model $M$ and the amount of available data for training the draft model $D$ and target model $D^\prime$.

\subsection{Motivation and Numerical Approximation Methodology}
\label{sec:motivation_optimal_N}

We aim to derive a simple, reusable rule linking the optimal draft model size \( N^\ast \) to the target model size \( M \) and dataset sizes of the draft model \( D \) and target model \(D^\prime\). While the ideal approach is to maximize throughput from Equation \eqref{eq:throughput_N} with respect to \( N \), the equation's complexity renders analytical optimization intractable.

Instead, we study Equation \eqref{eq:throughput_N} numerically, evaluating throughput over a dense mesh of \( (N,D, M,D^\prime) \) values. Parameters are sampled from logarithmically spaced ranges (Table~\ref{table:ranges}) to reflect realistic LLM scaling scenarios. For each \( (D, M, D^\prime) \), we identified the \( N \) value yielding the highest throughput as \( N^\ast \).

\begin{figure}[H]
    \centering
    \begin{subfigure}[b]{\linewidth}
        \centering
        \includegraphics[width=\linewidth]{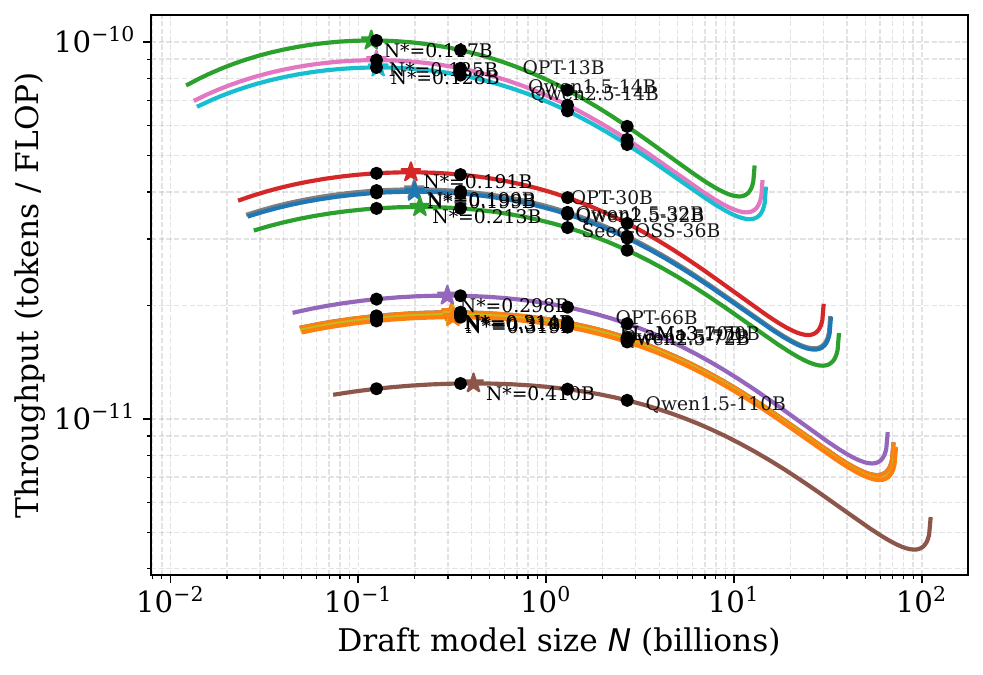}
        \caption{Throughput predicted by \eqref{eq:throughput_N} for the target models paired with OPT-based draft models.}
        \label{fig:throughput_curves_opt_draft}
    \end{subfigure}
    \\
    \begin{subfigure}[b]{\linewidth}
        \centering
        \includegraphics[width=\linewidth]{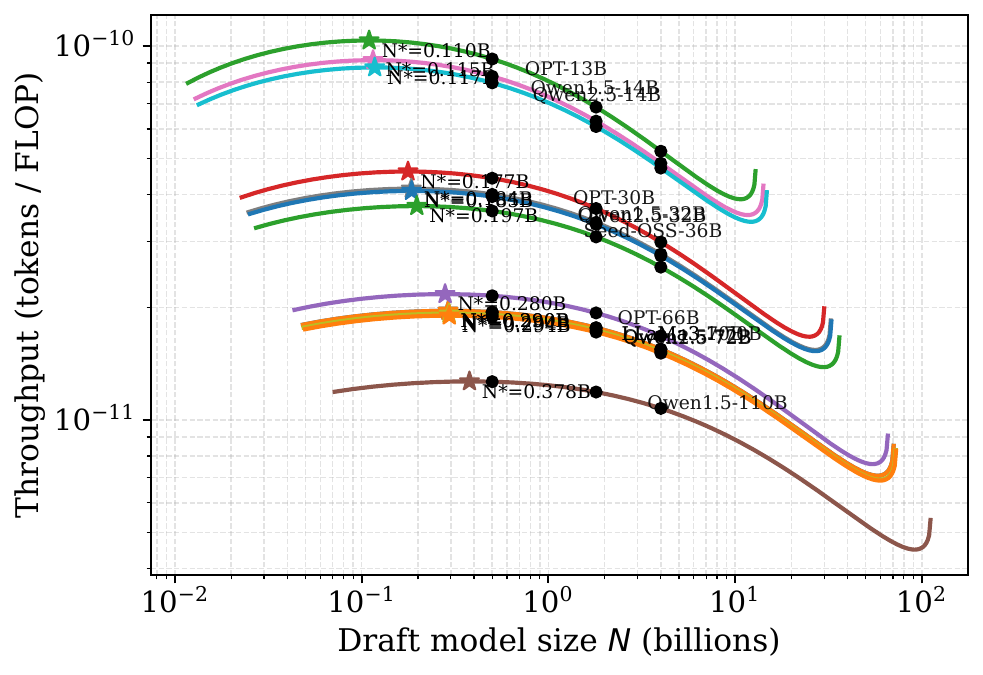} 
        \caption{Throughput predicted by \eqref{eq:throughput_N} for the target models paired with Qwen1.5-based draft models.}
        \label{fig:throughput_curves_qwen1.5_draft}
    \end{subfigure}
    \\
    \begin{subfigure}[b]{\linewidth}
        \centering
        \includegraphics[width=\linewidth]{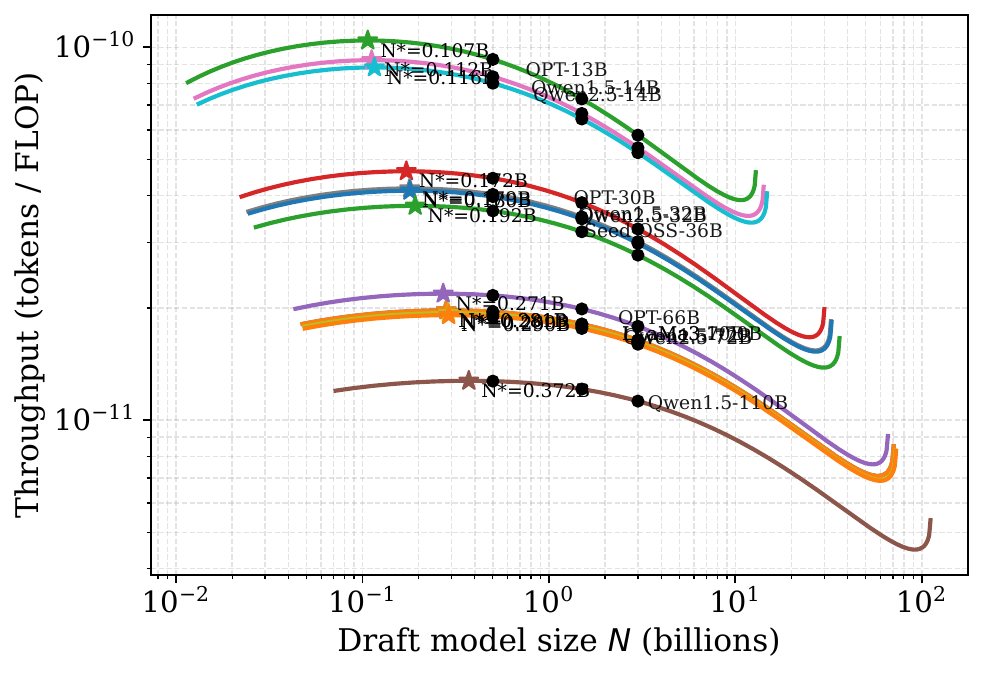} 
        \caption{Throughput predicted by \eqref{eq:throughput_N} for the target models paired with Qwen2.5-based draft models.}
        \label{fig:throughput_curves_qwen2.5_draft}
    \end{subfigure}
    
    \caption{Throughput (tokens per FLOP) predicted by Equation~\eqref{eq:throughput_N} as a function of draft model size \(N\) (in billions of parameters) for different target models and draft model families based on Table ~\ref{table:Optimal_draft_model_size_throughput}.
    Each curve corresponds to a single target model and is annotated directly on the curve.
    Black markers indicate the draft model sizes used in experiments, while star markers denote the optimal draft size \(N^\ast\) that maximizes predicted throughput for each target model.}
    \label{fig:combined_throughput}
\end{figure}

Across a broad range of target model sizes \(M\), draft training dataset sizes
\(D\), and target training dataset sizes \(D^\prime\), the dominant trend is a
strong and approximately linear growth of \(N^\ast\) with the target model size.
In contrast, variations in the training dataset sizes introduce only mild
modulations of this behavior.
These trends are illustrated in Figure~\ref{fig:Nstar_scaling_summary}.

\subsection{Basic Observation}
\label{Sec:observation_optimal_N}

The numerical analysis of Equation~\eqref{eq:throughput_N} reveals a clear and
systematic structure in the dependence of the throughput-optimal draft model
size \(N^\ast\) on the properties of the speculative decoding system.

In Figure~\ref{fig:Nstar_vs_M_raw}, the optimal draft size increases monotonically
with the target model size \(M\), and this trend is consistent across all
examined values of the draft training dataset size \(D\).
This behavior indicates that the dominant factor governing the magnitude of the
optimal draft model size is the target model size itself. The upward trend in \(N^\ast\) reflects the need for larger draft models as the target grows, while the downward trend in \(N^\ast/M\) indicates that this growth is approximately linear, with diminishing finite-size corrections.
In contrast, Figure~\ref{fig:Nstar_vs_D_raw} shows that varying the draft
training dataset size \(D\) at fixed \(M\) leads to only modest changes in
\(N^\ast\), suggesting that the influence of \(D\) is secondary.

To isolate these secondary effects, Figure~\ref{fig:Nstar_over_M_vs_M} and
Figure~\ref{fig:Nstar_over_M_vs_D} present the normalized quantity \(N^\ast / M\).
Normalization largely collapses the curves across different values of \(D\),
indicating that the primary scaling of the optimal draft size is approximately
linear in the target model size.
Residual variation in \(N^\ast / M\) across both \(M\) and \(D\) is mild and
systematic, reflecting sublinear corrections rather than a change in the
dominant scaling behavior.

Finally, Figures~\ref{fig:Nstar_vs_Dt_raw} and
\ref{fig:Nstar_over_M_vs_Dt} examine the sensitivity of the optimal
draft size to the target training dataset size \(D^\prime\).
Both the raw and normalized plots show that variations in \(D^\prime\) have a
negligible effect on \(N^\ast\) relative to the effects of \(M\) and \(D\),
indicating that target-side training data introduces only minor second-order
corrections to the draft model sizing rule.

\subsection{Analytical Approximation of the Throughput-Optimal Draft Size}
\label{sec:analytical_approx_nstar}

Motivated by the observed near-collapse of the normalized quantity \(N^\ast/M\) in Section \ref{Sec:observation_optimal_N}, we seek an analytical approximation that captures the leading scaling behavior while remaining interpretable and reusable. In particular, the numerical results suggest that \(N^\ast/M\) approaches a constant for large \(M\), with finite-size corrections that decay sub-linearly. This motivates the following scaling ansatz:
\begin{equation}
\frac{N^\ast}{M}
=
\mu
+
\frac{M_0}{M}
+
\gamma \log D
+
\gamma^\prime \log D^\prime,
\label{eq:nstar_scaling}
\end{equation}
where \(\mu\) represents the asymptotic draft-to-target size ratio, \(M_0\) captures finite-size corrections, and the coefficients \(\gamma\) and \(\gamma^\prime\) quantify residual dependence on the draft and target training dataset sizes, respectively.

We estimate the parameters in Equation~\eqref{eq:nstar_scaling} by fitting to the numerically computed optima \(N^\ast(M,D,D')\) obtained from exhaustive search over the draft size \(N\) for each configuration of \((M,D,D\prime)\). Each regression observation corresponds to one point in the Cartesian product of the logarithmically spaced grids in \(M\), \(D\), and \(D\prime\). The draft-size grid over \(N\) is used solely for the inner numerical optimization and does not contribute additional regression degrees of freedom. Ordinary least squares with heteroskedasticity-robust (HC3) standard errors is used to account for scale-dependent variance across model sizes.

The fitted parameters, reported in Table~\ref{table:nstar_scaling_moe_first}, confirm that the leading-order behavior of the throughput-optimal draft size is linear in the target model size. The asymptotic ratio \(\mu\) is tightly constrained, indicating that for sufficiently large target models the optimal draft is approximately constant in relative size. The finite-size term \(M_0/M\) explains the systematic decrease of \(N^\ast/M\) with increasing \(M\) observed in the numerical results.

Dataset-size effects enter only as small corrections. The coefficient \(\gamma\) is negative and statistically significant, indicating that increasing the amount of draft training data slightly reduces the optimal relative draft size, while the coefficient \(\gamma\prime\) associated with the target training dataset is statistically indistinguishable from zero. This confirms that target-side data availability has a negligible impact on draft sizing compared to the dominant dependence on \(M\).

\subsection{Summary}
\label{sec:optimal_draft_summary}

We analyzed how the throughput-optimal draft model size \(N^\ast\) depends on the target model size \(M\) and the training dataset sizes \((D,D\prime)\). The numerical results show that, across a wide range of configurations, the dominant factor governing the optimal draft size is the target model size itself, while dataset-related effects introduce only secondary corrections. Motivated by this observation, we derived a simplified leading-order scaling law that models \(N^\ast\) solely as a function of \(M\).

Using the full set of numerically computed optima across all combinations of \((M,D,D\prime)\), we performed a pooled linear regression to obtain a robust analytical approximation of the form
\begin{equation}
\label{eq:Nstar_vs_M}
N^\ast(M) = \mu M + M_0 .
\end{equation}
The fitted coefficients, along with their margins of error and confidence intervals, are reported in Table~\ref{table:nstar_linear_pooled}. This affine relation implies that the ratio \(N^\ast/M\) varies with target model size due to finite-size effects captured by the offset term \(M_0\).

For smaller target models, the additive constant \(M_0\) inflates the relative draft size, whereas its influence diminishes as \(M\) increases. In the large-\(M\) regime, the ratio \(N^\ast/M\) converges to the asymptotic value \(\mu \approx 2.7\times 10^{-3}\), indicating that the throughput-optimal draft model is approximately \(200\!\times\) smaller than the target. This convergence explains the systematic decrease of \(N^\ast/M\) observed in the numerical results and the near-collapse of normalized curves at large model sizes.

This asymptotic prediction provides a useful theoretical benchmark for speculative decoding. Empirical studies commonly explore draft-to-target size ratios spanning more than two orders of magnitude, with throughput-optimal configurations typically corresponding to draft models that are tens to hundreds of times smaller than the target. The scaling law derived here falls squarely within this empirically relevant regime, reinforcing its practical utility as a guideline for selecting draft model sizes in large-scale speculative decoding systems.

\section{Conclusion}

Speculative decoding accelerates large language model (LLM) inference by delegating token proposal to a lightweight draft model while relying on a larger target model for parallel verification. The effectiveness of this approach, however, is highly sensitive to the choice of the draft model, and suboptimal selections can significantly diminish throughput gains. Prior work has largely relied on empirical benchmarking to identify suitable draft models, resulting in increased deployment complexity and computational overhead.

In this work, we introduced \textbf{Speculative Decoding Scaling Laws (SDSL)}, an analytical framework that connects pre-training scaling laws to the throughput efficiency of speculative decoding systems. We showed that the expected token acceptance rate can be accurately modeled as an affine function of draft and target model perplexities, with draft model quality emerging as the dominant factor. By integrating this relationship with established pre-training scaling laws, we derived a principled, closed-form characterization of speculative decoding throughput in terms of model size and training data.

Our analysis reveals a robust and generalizable result: across model families, training regimes, and dataset scales, the throughput-optimal draft model size scales approximately linearly with the target model size, while dataset-related effects introduce only mild second-order corrections. In the large-model regime, this relationship converges to a constant draft-to-target size ratio on the order of $10^{-3}$, indicating that the optimal draft model is approximately $200\times$ smaller than the target. This prediction aligns closely with empirical observations across modern LLM families and provides a theoretical explanation for the regimes in which speculative decoding is most effective.

While our formulation is derived in terms of tokens per FLOP, we further validated its practical relevance through wall-clock latency measurements, demonstrating consistency across evaluation metrics. 

\section{Limitations}

Our framework assumes that both draft and target models are trained on comparable data distributions and with similar training recipes. Deviations from this assumption—such as heavy domain specialization, substantial architectural asymmetry, or aggressive post-training alignment—may affect the quantitative accuracy of the scaling coefficients, even if the qualitative trends remain valid.

In addition, our analysis focuses exclusively on autoregressive text-only language models. We do not consider alternative architectures such as encoder--decoder models, mixture-of-experts systems, or multi-modal models incorporating visual or audio inputs. Extending SDSL to these settings may require additional modeling assumptions and empirical validation.

\section*{Acknowledgements}
This work was supported by the National Science Foundation NRT-AI 2244574.

\newpage
\bibliography{custom}

@article{hellaswag,
  title={Hellaswag: Can a machine really finish your sentence?},
  author={Zellers, Rowan and Holtzman, Ari and Bisk, Yonatan and Farhadi, Ali and Choi, Yejin},
  journal={arXiv preprint arXiv:1905.07830},
  year={2019}
}

@inproceedings{xia2023speculative,
  title={Speculative Decoding: Exploiting Speculative Execution for Accelerating Seq2seq Generation},
  author={Xia, Heming and Ge, Tao and Wang, Peiyi and Chen, Si-Qing and Wei, Furu and Sui, Zhifang},
  booktitle={Findings of the Association for Computational Linguistics: EMNLP 2023},
  pages={3909--3925},
  year={2023}
}

@misc{DeepSpeed,
title={DeepSpeed},
author="Microsoft",
year=2023,
howpublished={\url{https://github.com/microsoft/deepspeed}},
urldate={2024-1-26},
note = "{Accessed: January 26, 2024}"
}

@article{yang2024multi,
  title={Multi-Candidate Speculative Decoding},
  author={Yang, Sen and Huang, Shujian and Dai, Xinyu and Chen, Jiajun},
  journal={arXiv preprint arXiv:2401.06706},
  year={2024}
}

@article{onlinespeculative,
  title={Online speculative decoding},
  author={Liu, Xiaoxuan and Hu, Lanxiang and Bailis, Peter and Stoica, Ion and Deng, Zhijie and Cheung, Alvin and Zhang, Hao},
  journal={arXiv preprint arXiv:2310.07177},
  year={2023}
}

@misc{leviathan2023fastinferencetransformersspeculative,
      title={Fast Inference from Transformers via Speculative Decoding}, 
      author={Yaniv Leviathan and Matan Kalman and Yossi Matias},
      year={2023},
      eprint={2211.17192},
      archivePrefix={arXiv},
      primaryClass={cs.LG},
      url={https://arxiv.org/abs/2211.17192}, 
}

@article{llama,
  title={Llama: Open and efficient foundation language models},
  author={Touvron, Hugo and Lavril, Thibaut and Izacard, Gautier and Martinet, Xavier and Lachaux, Marie-Anne and Lacroix, Timoth{\'e}e and Rozi{\`e}re, Baptiste and Goyal, Naman and Hambro, Eric and Azhar, Faisal and others},
  journal={arXiv preprint arXiv:2302.13971},
  year={2023}
}

@article{chen2023accelerating,
  title={Accelerating large language model decoding with speculative sampling},
  author={Chen, Charlie and Borgeaud, Sebastian and Irving, Geoffrey and Lespiau, Jean-Baptiste and Sifre, Laurent and Jumper, John},
  journal={arXiv preprint arXiv:2302.01318},
  year={2023}
}

@article{opt,
  title={Opt: Open pre-trained transformer language models},
  author={Zhang, Susan and Roller, Stephen and Goyal, Naman and Artetxe, Mikel and Chen, Moya and Chen, Shuohui and Dewan, Christopher and Diab, Mona and Li, Xian and Lin, Xi Victoria and others},
  journal={arXiv preprint arXiv:2205.01068},
  year={2022}
}

@article{zhang2023draft,
  title={Draft \& verify: Lossless large language model acceleration via self-speculative decoding},
  author={Zhang, Jun and Wang, Jue and Li, Huan and Shou, Lidan and Chen, Ke and Chen, Gang and Mehrotra, Sharad},
  journal={arXiv preprint arXiv:2309.08168},
  year={2023}
}

@misc{besiroglu2024chinchillascalingreplicationattempt,
      title={Chinchilla Scaling: A replication attempt}, 
      author={Tamay Besiroglu and Ege Erdil and Matthew Barnett and Josh You},
      year={2024},
      eprint={2404.10102},
      archivePrefix={arXiv},
      primaryClass={cs.AI},
      url={https://arxiv.org/abs/2404.10102}, 
}

@misc{hoffmann2022trainingcomputeoptimallargelanguage,
      title={Training Compute-Optimal Large Language Models}, 
      author={Jordan Hoffmann and Sebastian Borgeaud and Arthur Mensch and Elena Buchatskaya and Trevor Cai and Eliza Rutherford and Diego de Las Casas and Lisa Anne Hendricks and Johannes Welbl and Aidan Clark and Tom Hennigan and Eric Noland and Katie Millican and George van den Driessche and Bogdan Damoc and Aurelia Guy and Simon Osindero and Karen Simonyan and Erich Elsen and Jack W. Rae and Oriol Vinyals and Laurent Sifre},
      year={2022},
      eprint={2203.15556},
      archivePrefix={arXiv},
      primaryClass={cs.CL},
      url={https://arxiv.org/abs/2203.15556}, 
}

@misc{yan2024decodingspeculativedecoding,
      title={Decoding Speculative Decoding}, 
      author={Minghao Yan and Saurabh Agarwal and Shivaram Venkataraman},
      year={2024},
      eprint={2402.01528},
      archivePrefix={arXiv},
      primaryClass={cs.LG},
      url={https://arxiv.org/abs/2402.01528}, 
}

@misc{kaplan2020scalinglawsneurallanguage,
      title={Scaling Laws for Neural Language Models}, 
      author={Jared Kaplan and Sam McCandlish and Tom Henighan and Tom B. Brown and Benjamin Chess and Rewon Child and Scott Gray and Alec Radford and Jeffrey Wu and Dario Amodei},
      year={2020},
      eprint={2001.08361},
      archivePrefix={arXiv},
      primaryClass={cs.LG},
      url={https://arxiv.org/abs/2001.08361}, 
}

@misc{yan2025decodingspeculativedecoding,
      title={Decoding Speculative Decoding}, 
      author={Minghao Yan and Saurabh Agarwal and Shivaram Venkataraman},
      year={2025},
      eprint={2402.01528},
      archivePrefix={arXiv},
      primaryClass={cs.LG},
      url={https://arxiv.org/abs/2402.01528}, 
}

@misc{sun2024spectrfast,
      title={SpecTr: Fast Speculative Decoding via Optimal Transport}, 
      author={Ziteng Sun and Ananda Theertha Suresh and Jae Hun Ro and Ahmad Beirami and Himanshu Jain and Felix Yu},
      year={2024},
      eprint={2310.15141},
      archivePrefix={arXiv},
      primaryClass={cs.LG},
      url={https://arxiv.org/abs/2310.15141}, 
}

@inproceedings{
zhou2024distillspec,
title={DistillSpec: Improving Speculative Decoding via Knowledge Distillation},
author={Yongchao Zhou and Kaifeng Lyu and Ankit Singh Rawat and Aditya Krishna Menon and Afshin Rostamizadeh and Sanjiv Kumar and Jean-Fran{\c{c}}ois Kagy and Rishabh Agarwal},
booktitle={The Twelfth International Conference on Learning Representations},
year={2024},
url={https://openreview.net/forum?id=rsY6J3ZaTF}
}

@article{sun2024triforce,
  title={Triforce: Lossless acceleration of long sequence generation with hierarchical speculative decoding},
  author={Sun, Hanshi and Chen, Zhuoming and Yang, Xinyu and Tian, Yuandong and Chen, Beidi},
  journal={arXiv preprint arXiv:2404.11912},
  year={2024}
}

@article{chen2024magicdec,
  title={MagicDec: Breaking the Latency-Throughput Tradeoff for Long Context Generation with Speculative Decoding},
  author={Chen, Jian and Tiwari, Vashisth and Sadhukhan, Ranajoy and Chen, Zhuoming and Shi, Jinyuan and Yen, Ian En-Hsu and Chen, Beidi},
  journal={arXiv preprint arXiv:2408.11049},
  year={2024}
}

@ARTICLE{2024arXiv240721783G,
       author = {{Grattafiori}, Aaron and {Dubey}, Abhimanyu and {Jauhri}, Abhinav and {Pandey}, Abhinav and {Kadian}, Abhishek and {Al-Dahle}, Ahmad and {Letman}, Aiesha and {Mathur}, Akhil and {Schelten}, Alan and {Vaughan}, Alex and {Yang}, Amy and {Fan}, Angela and {Goyal}, Anirudh and {Hartshorn}, Anthony and {Yang}, Aobo and {Mitra}, Archi and {Sravankumar}, Archie and {Korenev}, Artem and {Hinsvark}, Arthur and {Rao}, Arun and {Zhang}, Aston and {Rodriguez}, Aurelien and {Gregerson}, Austen and {Spataru}, Ava and {Roziere}, Baptiste and {Biron}, Bethany and {Tang}, Binh and {Chern}, Bobbie and {Caucheteux}, Charlotte and {Nayak}, Chaya and {Bi}, Chloe and {Marra}, Chris and {McConnell}, Chris and {Keller}, Christian and {Touret}, Christophe and {Wu}, Chunyang and {Wong}, Corinne and {Canton Ferrer}, Cristian and {Nikolaidis}, Cyrus and {Allonsius}, Damien and {Song}, Daniel and {Pintz}, Danielle and {Livshits}, Danny and {Wyatt}, Danny and {Esiobu}, David and {Choudhary}, Dhruv and {Mahajan}, Dhruv and {Garcia-Olano}, Diego and {Perino}, Diego and {Hupkes}, Dieuwke and {Lakomkin}, Egor and {AlBadawy}, Ehab and {Lobanova}, Elina and {Dinan}, Emily and {Smith}, Eric Michael and {Radenovic}, Filip and {Guzm{\'a}n}, Francisco and {Zhang}, Frank and {Synnaeve}, Gabriel and {Lee}, Gabrielle and {Anderson}, Georgia Lewis and {Thattai}, Govind and {Nail}, Graeme and {Mialon}, Gregoire and {Pang}, Guan and {Cucurell}, Guillem and {Nguyen}, Hailey and {Korevaar}, Hannah and {Xu}, Hu and {Touvron}, Hugo and {Zarov}, Iliyan and {Arrieta Ibarra}, Imanol and {Kloumann}, Isabel and {Misra}, Ishan and {Evtimov}, Ivan and {Zhang}, Jack and {Copet}, Jade and {Lee}, Jaewon and {Geffert}, Jan and {Vranes}, Jana and {Park}, Jason and {Mahadeokar}, Jay and {Shah}, Jeet and {van der Linde}, Jelmer and {Billock}, Jennifer and {Hong}, Jenny and {Lee}, Jenya and {Fu}, Jeremy and {Chi}, Jianfeng and {Huang}, Jianyu and {Liu}, Jiawen and {Wang}, Jie and {Yu}, Jiecao and {Bitton}, Joanna and {Spisak}, Joe and {Park}, Jongsoo and {Rocca}, Joseph and {Johnstun}, Joshua and {Saxe}, Joshua and {Jia}, Junteng and {Vasuden Alwala}, Kalyan and {Prasad}, Karthik and {Upasani}, Kartikeya and {Plawiak}, Kate and {Li}, Ke and {Heafield}, Kenneth and {Stone}, Kevin and {El-Arini}, Khalid and {Iyer}, Krithika and {Malik}, Kshitiz and {Chiu}, Kuenley and {Bhalla}, Kunal and {Lakhotia}, Kushal and {Rantala-Yeary}, Lauren and {van der Maaten}, Laurens and {Chen}, Lawrence and {Tan}, Liang and {Jenkins}, Liz and {Martin}, Louis and {Madaan}, Lovish and {Malo}, Lubo and {Blecher}, Lukas and {Landzaat}, Lukas and {de Oliveira}, Luke and {Muzzi}, Madeline and {Pasupuleti}, Mahesh and {Singh}, Mannat and {Paluri}, Manohar and {Kardas}, Marcin and {Tsimpoukelli}, Maria and {Oldham}, Mathew and {Rita}, Mathieu and {Pavlova}, Maya and {Kambadur}, Melanie and {Lewis}, Mike and {Si}, Min and {Singh}, Mitesh Kumar and {Hassan}, Mona and {Goyal}, Naman and {Torabi}, Narjes and {Bashlykov}, Nikolay and {Bogoychev}, Nikolay and {Chatterji}, Niladri and {Zhang}, Ning and {Duchenne}, Olivier and {{\c{C}}elebi}, Onur and {Alrassy}, Patrick and {Zhang}, Pengchuan and {Li}, Pengwei and {Vasic}, Petar and {Weng}, Peter and {Bhargava}, Prajjwal and {Dubal}, Pratik and {Krishnan}, Praveen and {Singh Koura}, Punit and {Xu}, Puxin and {He}, Qing and {Dong}, Qingxiao and {Srinivasan}, Ragavan and {Ganapathy}, Raj and {Calderer}, Ramon and {Silveira Cabral}, Ricardo and {Stojnic}, Robert and {Raileanu}, Roberta and {Maheswari}, Rohan and {Girdhar}, Rohit and {Patel}, Rohit and {Sauvestre}, Romain and {Polidoro}, Ronnie and {Sumbaly}, Roshan and {Taylor}, Ross and {Silva}, Ruan and {Hou}, Rui and {Wang}, Rui and {Hosseini}, Saghar and {Chennabasappa}, Sahana and {Singh}, Sanjay and {Bell}, Sean and {Kim}, Seohyun Sonia and {Edunov}, Sergey and {Nie}, Shaoliang and {Narang}, Sharan and {Raparthy}, Sharath and {Shen}, Sheng and {Wan}, Shengye and {Bhosale}, Shruti and {Zhang}, Shun and {Vandenhende}, Simon and {Batra}, Soumya and {Whitman}, Spencer and {Sootla}, Sten and {Collot}, Stephane and {Gururangan}, Suchin and {Borodinsky}, Sydney and {Herman}, Tamar and {Fowler}, Tara and {Sheasha}, Tarek and {Georgiou}, Thomas and {Scialom}, Thomas and {Speckbacher}, Tobias},
        title = "{The Llama 3 Herd of Models}",
      journal = {arXiv e-prints},
         year = 2024,
        month = jul,
          eid = {arXiv:2407.21783},
        pages = {arXiv:2407.21783},
          doi = {10.48550/arXiv.2407.21783},
archivePrefix = {arXiv},
       eprint = {2407.21783},
 primaryClass = {cs.AI},
       adsurl = {https://ui.adsabs.harvard.edu/abs/2024arXiv240721783G}
}

@misc{seed2025seed-oss,
  author={ByteDance Seed Team},
  title={Seed-OSS Open-Source Models},
  year={2025},
  howpublished={\url{https://github.com/ByteDance-Seed/seed-oss}}
}

@misc{qwen2.5,
    title = {Qwen2.5: A Party of Foundation Models},
    url = {https://qwenlm.github.io/blog/qwen2.5/},
    author = {Qwen Team},
    month = {September},
    year = {2024}
}

@article{qwen,
  title={Qwen Technical Report},
  author={Jinze Bai and Shuai Bai and Yunfei Chu and Zeyu Cui and Kai Dang and Xiaodong Deng and Yang Fan and Wenbin Ge and Yu Han and Fei Huang and Binyuan Hui and Luo Ji and Mei Li and Junyang Lin and Runji Lin and Dayiheng Liu and Gao Liu and Chengqiang Lu and Keming Lu and Jianxin Ma and Rui Men and Xingzhang Ren and Xuancheng Ren and Chuanqi Tan and Sinan Tan and Jianhong Tu and Peng Wang and Shijie Wang and Wei Wang and Shengguang Wu and Benfeng Xu and Jin Xu and An Yang and Hao Yang and Jian Yang and Shusheng Yang and Yang Yao and Bowen Yu and Hongyi Yuan and Zheng Yuan and Jianwei Zhang and Xingxuan Zhang and Yichang Zhang and Zhenru Zhang and Chang Zhou and Jingren Zhou and Xiaohuan Zhou and Tianhang Zhu},
  journal={arXiv preprint arXiv:2309.16609},
  year={2023}
}

@misc{meta_llama3_1,
  title={{Introducing Llama 3.1: Our most capable models to date}},
  author={Meta AI},
  year={2024},
  howpublished={\url{https://ai.meta.com/blog/meta-llama-3-1/}}
}

@misc{gu2025dartdenoisingautoregressivetransformer,
      title={DART: Denoising Autoregressive Transformer for Scalable Text-to-Image Generation}, 
      author={Jiatao Gu and Yuyang Wang and Yizhe Zhang and Qihang Zhang and Dinghuai Zhang and Navdeep Jaitly and Josh Susskind and Shuangfei Zhai},
      year={2025},
      eprint={2410.08159},
      archivePrefix={arXiv},
      primaryClass={cs.CV},
      url={https://arxiv.org/abs/2410.08159}, 
}
\newpage
\appendix

\section{Related Works} \label{sec:related}

Optimizing draft model design has been a key area of research for improving speculative decoding efficiency. \citet{yan2025decodingspeculativedecoding} conducted a comprehensive benchmarking study to investigate the factors that influence throughput improvements in speculative decoding.  They identified \textbf{draft model latency} as the primary bottleneck, highlighting that deeper models with the same parameter count exhibit higher latency. Furthermore, they observed that \textbf{accuracy on language modeling tasks} does not strongly correlate with speculative decoding performance, suggesting that choosing a draft model based solely on accuracy may be suboptimal for maximizing throughput. To address these issues, the authors propose redesigning draft models by adjusting their depth-to-width ratio to optimize throughput, resulting in up to a 60\% improvement in efficiency. Their work emphasizes the need for a systematic approach to draft model design and shows how these improvements can lead to substantial performance gains, such as reducing KV-cache requirements by 37\%, enabling larger batch sizes, and outperforming other methods like self-speculative decoding \cite{zhang2023draft}.

~\citet{onlinespeculative} proposed continuously training the draft model on the outputs of the target model to improve token acceptance rates, though performing such training during inference remains challenging. ~\citet{xia2023speculative} introduced an encoder-decoder-based draft model, providing an alternative to traditional autoregressive architectures. Another line of work focuses on increasing the number of candidate tokens per step, as investigated by ~\citet{sun2024spectrfast} and ~\citet{yang2024multi}, which improves throughput by enhancing the acceptance rate. Additionally, ~\citet{zhou2024distillspec} explored draft model distillation to create more efficient and compact models, aligning with the goal of reducing computational overhead while maintaining performance. In the context of long-context scenarios, ~\citet{sun2024triforce} and ~\citet{chen2024magicdec} examined speculative decoding strategies tailored for extended sequences, reinforcing the need for systematic draft model design. 

~\citet{gu2025dartdenoisingautoregressivetransformer} introduced DART, which integrates autoregressive and diffusion processes within a non-Markovian framework. By eliminating the Markovian assumption that limits traditional diffusion models, DART leverages the full generative trajectory, allowing for more efficient image generation. This approach iteratively denoises image patches using an autoregressive model, enhancing the quality of text-to-image generation. Additionally, DART removes the need for image quantization, significantly improving computational efficiency while maintaining flexibility, ultimately setting a new standard for high-quality, scalable image synthesis.

The LIMINAL model \cite{gu2025dartdenoisingautoregressivetransformer} provides an analytical framework to assess the fundamental performance limits of Large Language Model (LLM) inference, with a particular focus on the auto-regressive decoding phase. It abstracts hardware and application parameters—such as compute throughput, memory bandwidth, memory capacity, and inter-chip communication latency—enabling the evaluation of system performance across a wide range of existing and projected hardware configurations, including GPUs and TPUs. The model reveals that the performance of LLM inference is primarily constrained by memory bandwidth, synchronization latency, and compute capacity. 

\section{Estimation of Parameter \(\alpha\)}
\label{sec:alpha_CI}

To estimate the value of $\alpha$ characterizing a specific pair of draft ($M_q$) and target ($M_p$) models, we first estimate the Token Acceptance Rate (TAR) between them. TAR quantifies the probability that a draft token generated by $M_q$ is accepted by $M_p$ during speculative decoding, and thus serves as a key observable reflecting the relationship between the two models.

We measure TAR as a function of the lookahead length $\gamma$ using the procedure described in Appendix~\ref{sec:TAR}. Once the TAR values are collected, we use them to estimate $\alpha$ by solving an inverse problem: specifically, we fit the observed TARs to the theoretical model given by \eqref{eq:TAR_α}. This fitting involves solving a nonlinear, overdetermined system of nine equations—one for each value of $\gamma$—using a least-squares formulation to minimize residuals. we also quantify the uncertainty of the estimated $\alpha$ by constructing confidence intervals, as detailed in Appendix~\ref{sec:CI}.

\subsection{Tokens Accepted Rates}
\label{sec:TAR}

The \textbf{Token Acceptance Rate (TAR)} introduced by \citet{yan2025decodingspeculativedecoding} serves as a crucial metric in evaluating the performance of speculative decoding algorithms. Defined as the average rate at which tokens generated by the draft model \( M_q \) are accepted by the target model \( M_p \), TAR quantifies the effectiveness of the approximation made by \( M_q \). Empirically, TAR can be achieved by dividing the number of accepted tokens by the number of tokens generated by the draft model \( M_q \) during decoding. For a given \(\gamma\), also called lookahead length representing the maximum number of tokens generated by the draft model, TAR can also be calculated using the formula suggested by \citet{leviathan2023fastinferencetransformersspeculative}:

\begin{equation}
    \label{eq:TAR_α}
    \text{TAR}(\gamma) = \frac{1 - \alpha^{\gamma+1}}{1 - \alpha}
\end{equation}

This equation illustrates how TAR is influenced by the expected acceptance rate \(\alpha\) and highlights how well \( M_q \) approximates \( M_p \). A higher acceptance rate indicates more efficient sampling, leading to an increased number of tokens produced with each run \citep{yan2025decodingspeculativedecoding}.

The draft model \( M_q \) samples \( x_{1,\ldots,\gamma} \) guesses in an autoregressive manner. Specifically, for each guess \( i \) (where \( i \) ranges from 1 to \(\gamma\)), the probability distribution is computed as follows:
$$
q_i(x) = M_q(\text{prefix} + [x_1, \ldots, x_{i-1}])
$$

Then, a token \( x_i \) is sampled from the distribution \( q_i(x) \), corresponding to the probability of the next token given the prefix extended by previously sampled guesses.
After generating all guesses, the target model \( M_p \) computes the probability of each guess by evaluating the extended prefixes \((\text{prefix} + x_1), (\text{prefix} + x_1, x_2), \ldots, (\text{prefix} + x_1, \ldots, x_\gamma)\) in parallel. 

The number of accepted guesses \( n \) is then determined by comparing random samples drawn from a uniform distribution against the ratio of probabilities from the target and draft models:
$$
n = \min(\{ i - 1 \mid 1 \le i \le \gamma, r_i > \frac{p_i(x)}{q_i(x)}\} \cup \{\gamma\})
$$

Where \( r_i \sim U(0,1) \) denotes a random sample drawn independently from the uniform distribution for each guess \( i \) .This process, repeated across multiple speculative decoding steps, enables empirical measurement of TAR at different lookahead lengths. 


After repeating the speculative decoding steps for a fixed \(\gamma\) across multiple trials, the empirical TAR is computed as the mean number of accepted tokens divided by the number of tokens proposed:

$$
\text{TAR}_{\gamma} = \text{mean}(n)
$$


This metric provides a practical measure of the average acceptance rate under different lookahead settings, characterizing the efficiency of speculative decoding.

\subsection{Calculating confidence intervals for $\alpha$ estimation}\label{sec:CI}

Let \(b_i\) denote the empirically observed Token Acceptance Rate (TAR) for the specific pair of target and draft models and the corresponding lookahead length \(\gamma_i\). 

\begin{itemize}
    	
         \item \textbf{\(E(\alpha, \gamma_i)\)}: Represents the predicted value from the model function for a given parameter \(\alpha\) and a specific power \(\gamma_i\), calculating an expected outcome based on the model, which is theoretical in nature.

    \end{itemize}

    The goal of estimating \(\alpha\) is to minimize the difference between these two quantities.

    \paragraph{Extracting the Residuals and Jacobian}
    
    After performing the nonlinear least squares fitting, we extract two important components from the optimization result,  the residuals and the Jacobian matrix : 
    
    \begin{enumerate}

    \item \textbf{Residuals}
    denoted as \( r_i(\alpha) \), represent the differences between the predicted values from our model function \( E(\alpha, \gamma_i) \) and the observed TARs \( b_i \). They are defined as:
    $$
    r_i(\alpha) = E(\alpha, \gamma_i) - b_i
    $$
    
    The collection of all residuals can be expressed in vector form as:
    $$
    \mathbf{r} = 
    \begin{bmatrix}
    	r_1(\alpha) \\ 
    	r_2(\alpha) \\ 
    	\vdots \\ 
    	r_n(\alpha) 
    \end{bmatrix} = 
    \begin{bmatrix}
    	E(\alpha, \gamma_1) - b_1 \\ 
    	E(\alpha, \gamma_2) - b_2 \\ 
    	\vdots \\ 
    	E(\alpha, \gamma_n) - b_n 
    \end{bmatrix}
    $$
     This residual quantifies how far off your model's prediction is from the actual observed TAR. By minimizing these residuals across all data pairs, we can estimate an optimal value for \(\alpha\).
     
     	\item \textbf{Jacobian Matrix}, denoted as \( J \), contains the first derivatives of the residuals with respect to the parameter \( \alpha \). Mathematically, if we denote our residuals as \( r_i(\alpha) \) for each data pair, then the Jacobian is defined as:
    $$
    J_{i,j} = \frac{\partial r_i(\alpha)}{\partial \alpha}
    $$
    This matrix provides information about how sensitive the residuals are to changes in the parameter \( \alpha \). In other words, it indicates how small changes in \( \alpha \) will affect the residuals.The Jacobian is crucial for optimization algorithms because it helps determine the direction and magnitude of updates to the parameter during the fitting process. 
 
    \end{enumerate}

       \paragraph{Variance of Residuals}
       
       To estimate the variance of the residuals, we first need to define the total number of observed TARs, denoted as \( n_{\text{\(b_i\)}} \). The degrees of freedom (dof) are defined as:
       
       $$
       \text{dof} = n_{\text{\(b_i\)}} - 1
       $$

       The estimated variance of the residuals is calculated under the assumption that they follow a normal distribution. This is computed using the formula:
       
       $$
       \sigma^2_{\text{residual}} = \frac{\sum_{i=1}^{n_{\text{\(b_i\)}}} r_i^2(\alpha)}{\text{dof}}
       $$

       In this equation, \( r_i(\alpha) \) represents the residual for each data pair, By summing the squared residuals and dividing by the degrees of freedom, we obtain an unbiased estimate of the variance of the residuals.
     \paragraph{Variance of \( \alpha \), Standard Error, and Confidence Intervals}
     
     The variance of the estimate for parameter \( \alpha \) is given by:
     
     $$
     \text{Var}(\alpha) = \sigma^2_{\text{residual}} (J^T J)^{-1}
     $$

     The standard deviation of \( \alpha \) is then calculated as:
     
     $$
     \text{Std}(\alpha) = \sqrt{\text{Var}(\alpha)}
     $$

     To construct a 95\% confidence interval for the estimated parameter \( \alpha \), we utilize the standard deviation of the estimate, denoted as \( \text{Std}(\alpha) \).
     
     The confidence interval can be expressed as:
     
     $$
     \text{Lower Bound} = {\alpha} - z_{0.025} \cdot \text{Std}({\alpha})
     $$

     $$
     \text{Upper Bound} = {\alpha} + z_{0.025} \cdot \text{Std}({\alpha})
     $$

     Where:
     \begin{itemize}
     	\item \( z_{0.025} \) is the critical value from the standard normal distribution corresponding to a 95\% confidence level (approximately 1.96).
     \end{itemize}
     
     This methodology allows us to quantify the uncertainty around our estimate of \( \alpha \), providing a range within which we can be 95\% confident that the true parameter value lies.

\section{Optimization of Throughput over $\gamma$}\label{sec:gamma_derivation}

We defined \(\gamma_{\text{optimal}}\) as the optimal number of tokens that the draft model should generate to maximize throughput. We derived this value by solving the derivative of Equation \eqref{eq:Throughput_orginal} with respect to \(\gamma\):
{
\[
\frac{d \mathcal{T}}{d\gamma} = \frac{N - N \alpha^{1 + \gamma} + (M + N \gamma) \alpha^{1 + \gamma} \log(\alpha)}{2 (M + N \gamma)^2 (-1 + \alpha)}
\]
}

To find the optimal \( \gamma \), we check where this derivative equals zero, leading to:
{
\begin{equation}
\label{eq:gamma_optimal}
\gamma_{\text{opt}} = \frac{-M \log(\alpha) + N \, W\left(-\frac{\alpha^{(M/N - 1)}}{e}\right) + N}{N \log(\alpha)}
\end{equation}
}

Substituting \eqref{eq:gamma_optimal} into \eqref{eq:Throughput_orginal} yields the throughput at optimal \(\gamma\), shown in \eqref{eq:Throughput}.


Furthermore, by incorporating \eqref{eq:plane} and \eqref{eq:Perplexity} into the optimality condition in \eqref{eq:gamma_optimal}, we derive an explicit expression for \(\gamma_{\text{optimal}}\) in terms of the draft and target model parameters:

\begin{align}
\label{eq:γ_optimal}
\gamma_{\text{opt}} &= \frac{-M \log(Ae^{L(N, D)} + Be^{L(M,D^\prime)} + C)}{N \log(Ae^{L(N, D)} + Be^{L(M,D^\prime)} + C)} \nonumber \\
&\quad + \frac{N \cdot W\left(-\frac{(Ae^{L(N, D)} + Be^{L(M,D^\prime)} + C)^{(M/N - 1)}}{e} \right)}{N \log(Ae^{L(N, D)} + Be^{L(M,D^\prime)} + C)} \nonumber \\
&\quad + \frac{N}{N \log(Ae^{L(N, D)} + Be^{L(M,D^\prime)} + C)}
\end{align}


where \( W \) denotes the Lambert W function. Since the argument inside \( W(x) \) is negative, we specifically use the \textbf{\( W_{-1} \)} branch, which provides real solutions in the range \( -1/e \leq x < 0 \). This selection ensures that \( \gamma_{\text{optimal}} \) remains well-defined and avoids complex-valued results. However, since \( \alpha \) is always in the range \( (0,1) \) and \( M \) is typically much larger than \( N \), the argument of the Lambert W function, \( -\frac{\alpha^{(M/N - 1)}}{e} \), remains within the valid domain of the \( W_{-1} \) branch, ensuring real-valued solutions. This guarantees the applicability of the derived expression for \( \gamma_{\text{optimal}} \) under standard conditions. The model is not applicable when \( M < N \).

Note that it is possible to implement a non-integer value of $\gamma$ by randomizing the choice of the lookahead length between the steps of the speculative decoding procedure. 
This formula depends on the key architecture hyperparameters $M$ and $N$ of speculative decoding system components, along with $\alpha,$ which captures the alignment between the components.

\section{Regressing \boldmath{$\alpha$} on perplexity of draft models}
\label{sec:regression_on_draft_model}

\begin{figure}[ht]
    \centering
    \begin{subfigure}[b]{\linewidth}
        \centering
        \includegraphics[width=\linewidth]{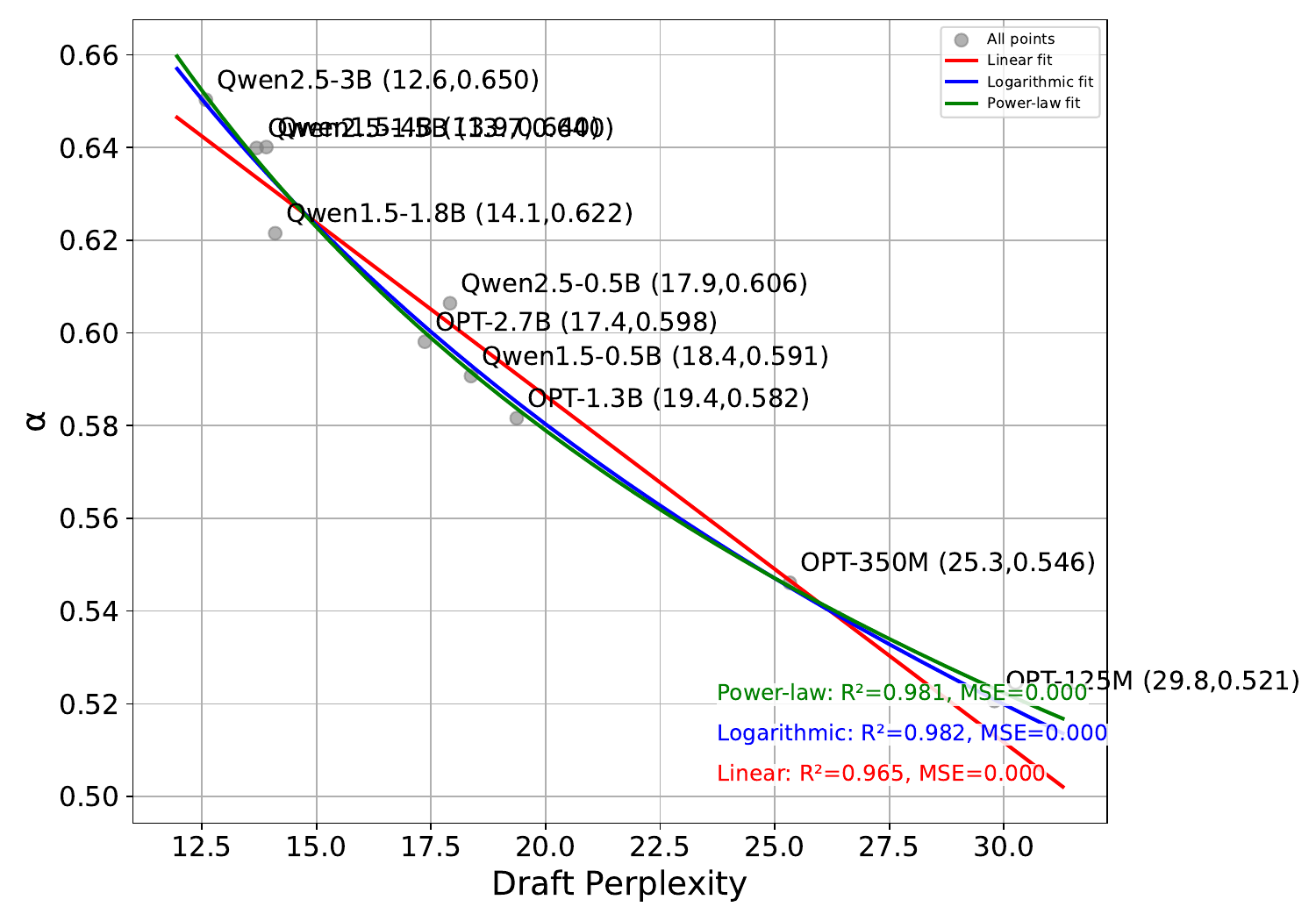}
        \caption{Curves relating the estimated $\alpha$ to the perplexity of the draft models paired with the Qwen1.5-110B target model.}
        \label{fig:qwen1.5_110B}
    \end{subfigure}
    
    \vspace{0.25cm} 
    
    \begin{subfigure}[b]{\linewidth}
        \centering
        \includegraphics[width=\linewidth]{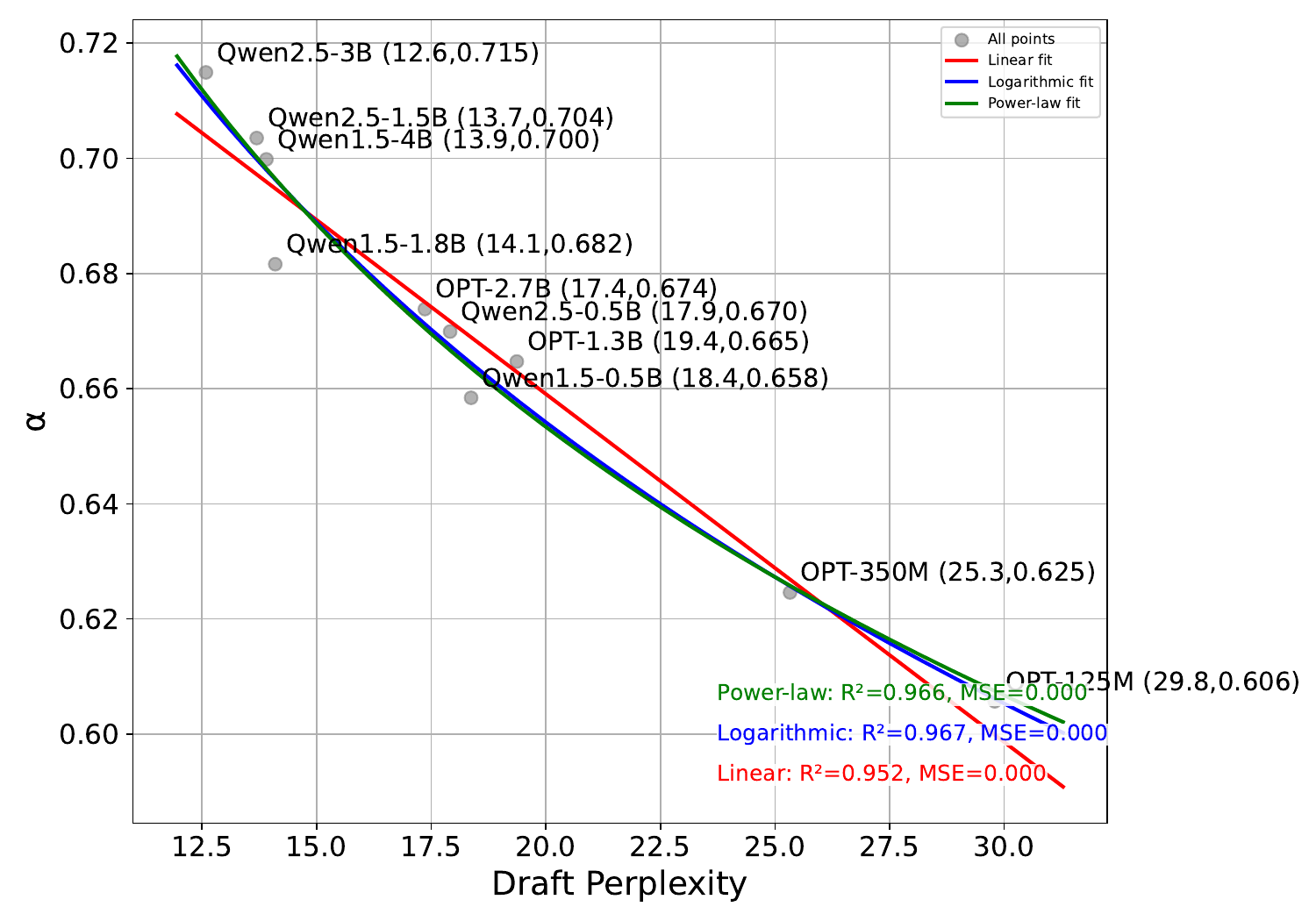} 
        \caption{Curves relating the estimated $\alpha$ to the perplexity of the draft models paired with the LLaMA3.1-70B target model.} 
        \label{fig:LLaMA3.1_70B}
    \end{subfigure}
    \caption{Curves show estimated \(\alpha\) values as a function of draft model perplexity, based on data from Tables~\ref{Table:Alpha_perplexity:OPT_drafts} and~\ref{Table:Alpha_perplexity:Qwen_drafts}.  The \(\alpha\) values were computed based on the method outlined in Appendix~\ref{sec:alpha_CI}, which estimates them using accepted token statistics. Function details appear in Table~\ref{table:alpha_regression_draft}.
}
    \label{fig:alpha_regression_draft}
\end{figure}

 To analyze the relationship between draft model perplexity ($x$) and the estimated scaling parameter $\alpha$, we fitted a set of analytical scaling laws—including power-law, linear, and logarithmic functions—to the data presented in Tables~\ref{Table:Alpha_perplexity:OPT_drafts} and~\ref{Table:Alpha_perplexity:Qwen_drafts}. For each target model, the fits were performed using paired values of draft model perplexity and the corresponding estimated $\alpha$ across multiple draft models. We restrict our analysis to scaling laws with two free parameters, reflecting the limited number of available draft models and, consequently, the small number of data points per target model. The estimated parameters of the fitted scaling laws, together with their margins of error, 95\% confidence intervals, mean squared error (MSE), and coefficient of determination ($R^2$), are summarized in Table~\ref{table:alpha_regression_draft} for all target models considered.
 The fitted relationships are illustrated in Figures \ref{fig:qwen1.5_110B} and \ref{fig:LLaMA3.1_70B}.

\paragraph{Results.}
Table~\ref{table:alpha_regression_draft} and Figure~\ref{fig:alpha_regression_draft} consistently indicate a monotonic relationship between draft model perplexity and the estimated scaling parameter $\alpha$ across all target models and fitted functional forms. In particular, as the perplexity of the draft models decreases, the estimated values of $\alpha$ systematically increase, independent of the specific analytical scaling law employed. This trend is observed for linear, logarithmic, and power-law fits, and is consistent across the full range of target models considered.

\section{Regressing \boldmath{$\alpha$} on perplexity of target models}
\label{sec:regression_on_target_model}

Tables~\ref{Table:Alpha_perplexity:OPT_drafts} and~\ref{Table:Alpha_perplexity:Qwen_drafts} summarize the estimated $\alpha$ values obtained for different combinations of target models and draft models, together with the corresponding perplexities of the target models. Across both tables, $\alpha$ varies only weakly with changes in target model perplexity when the draft model is held fixed. In particular, for a given draft model, target models spanning a broad range of perplexity values often yield similar $\alpha$ estimates, with no consistent monotonic trend as a function of target model perplexity.

This behavior contrasts with the strong and systematic dependence of $\alpha$ on draft model perplexity observed in Section~\ref{sec:regression_on_draft_model}. While higher-quality target models (i.e., those with lower perplexity) sometimes correspond to slightly larger $\alpha$ values, this effect is neither uniform across draft models nor consistent across target model families. In several cases, target models with comparable perplexity produce nearly identical $\alpha$ estimates, despite substantial differences in their absolute perplexity values.

Overall, the results indicate that $\alpha$ is significantly more sensitive to the perplexity of the draft model than to that of the target model. At the scale of the OPT, Qwen, LLaMa, and Seed target models considered here, we do not observe a robust, quantifiable dependency of $\alpha$ on target model perplexity.

\section{Latency-based validation of \(N^\ast\) for OPT-13B}
\label{sec:validate_optimal_N}

In Section~\ref{sec:Throughput_Optimal_N}, we analytically derived the optimal draft model size \(N^\ast\) that maximizes the predicted throughput for each target model and draft family by maximizing Equation~\eqref{eq:throughput_N}. These predictions are summarized in Table~\ref{table:Optimal_draft_model_size_throughput}. While the analytical formulation captures the dominant computational trade-offs of speculative decoding, it abstracts away system-level effects that can influence real-world inference performance, including kernel launch overheads, memory bandwidth constraints, and model execution overheads. In this subsection, we empirically validate the predicted optimal draft size for the OPT-13B target model by directly measuring speculative decoding latency across all evaluated draft models.

\paragraph{Method.}
Latency measurements are performed on the HellaSwag dataset. We randomly select 50 prompts and use each prompt independently to measure inference latency. Prompts are tokenized using the target model tokenizer and truncated or padded as needed to ensure compatibility when draft and target models employ different tokenizers.

For each prompt, we measure three latency components: (i) \emph{time-to-first-token (TTFT)}, defined as the time required to generate the first output token under speculative decoding; (ii) \emph{total generation time (TTOT)}, measured for generating a fixed-length continuation of 250 tokens; and (iii) \emph{time-per-output-token (TPOT)}, computed as TTOT divided by the number of generated tokens. Speculative decoding is executed with greedy decoding (no sampling), a maximum generation length of 250 tokens, and batch size one. For draft–target pairs that do not share a tokenizer, both the draft and target tokenizers are explicitly provided during generation to ensure correct token alignment.

All models are loaded in half precision (\texttt{float16}) and executed on a single A100 GPU. To mitigate one-time initialization effects such as kernel compilation and cache warming, a warm-up phase consisting of a forward pass and a short speculative generation is performed prior to measurement. GPU synchronization is enforced before and after each timed segment to ensure accurate wall-clock measurements. For each draft model, latency metrics are aggregated across prompts, and we report the mean along with a 95\% confidence interval computed using the standard normal approximation. The resulting measurements are summarized in Table~\ref{table:latency_validation_opt13b}.

\paragraph{Results and validation.}
We emphasize that this appendix focuses on OPT-13B as a representative target model, and that absolute latency values may differ across draft families due to architectural and tokenizer differences though the similar trend is expected .
Table~\ref{table:latency_validation_opt13b} shows that, across all evaluated draft model families, the minimum observed latency consistently occurs at draft sizes that closely align with the analytically predicted optimal draft size \(N^\ast\). Draft models whose sizes lie closest to \(N^\ast\) achieve lower TTFT, TTOT, and TPOT compared to both substantially smaller and larger draft models.

 This agreement provides empirical support for the throughput-based optimization derived in Section~\ref{sec:Throughput_Optimal_N}, demonstrating that maximizing predicted throughput is an effective proxy for minimizing end-to-end inference latency. To quantify deviations from the predicted optimum, we report the normalized distance \(|N - N^\ast|/M\), which measures how far each evaluated draft size deviates from the analytical optimum relative to the target model size. Across all draft families, latency increases monotonically with this normalized distance, indicating that \(N^\ast\) accurately captures the location of the latency minimum even when only a discrete set of draft sizes is available.

\begin{figure}[H]
    \centering
    \begin{subfigure}[b]{\linewidth}
        \centering
        \includegraphics[width=\linewidth]{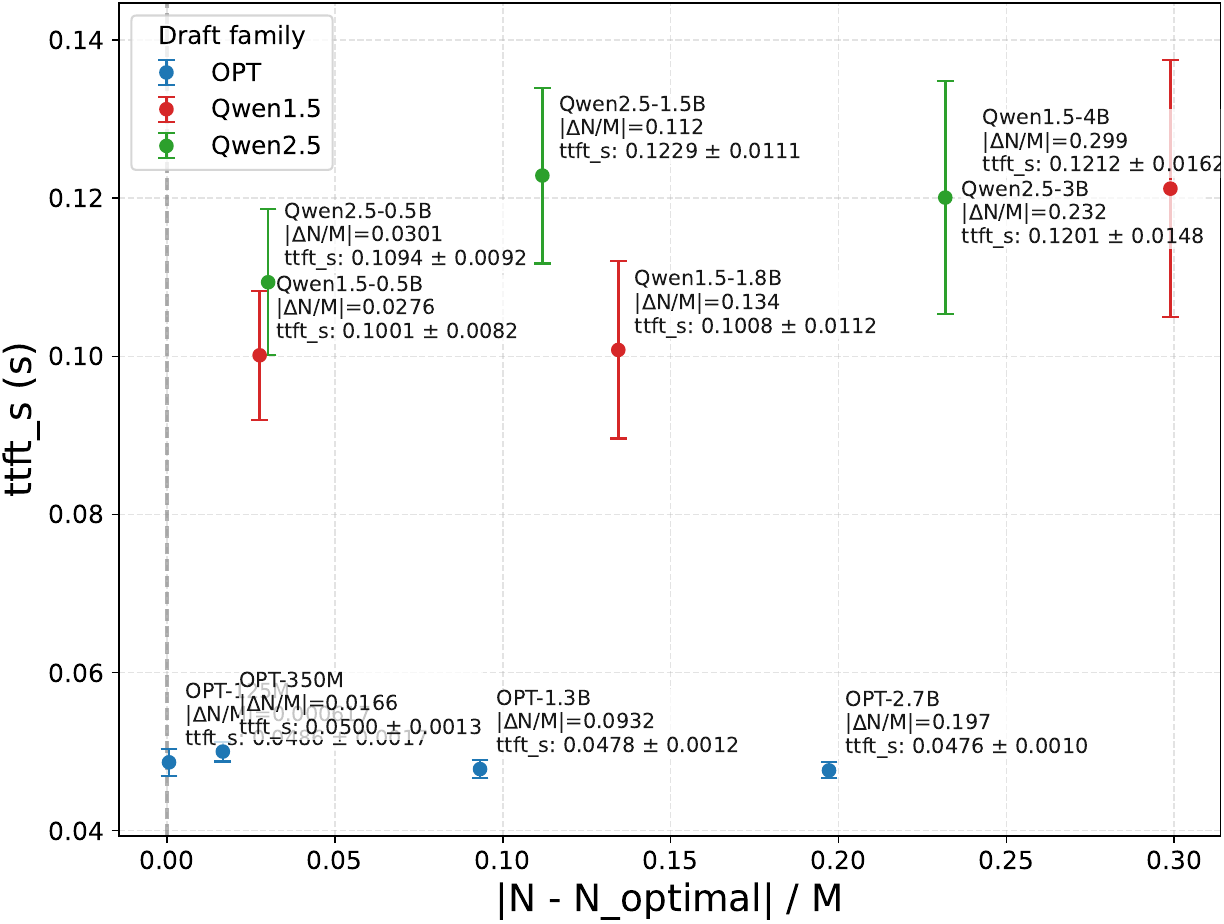}
        \caption{Measured time-to-first-token (TTFT) under speculative decoding for OPT-13B as a function of draft model size \(N\).}
        \label{fig:OPT_13B_TTFT}
    \end{subfigure}
    \\
    \begin{subfigure}[b]{\linewidth}
        \centering
        \includegraphics[width=\linewidth]{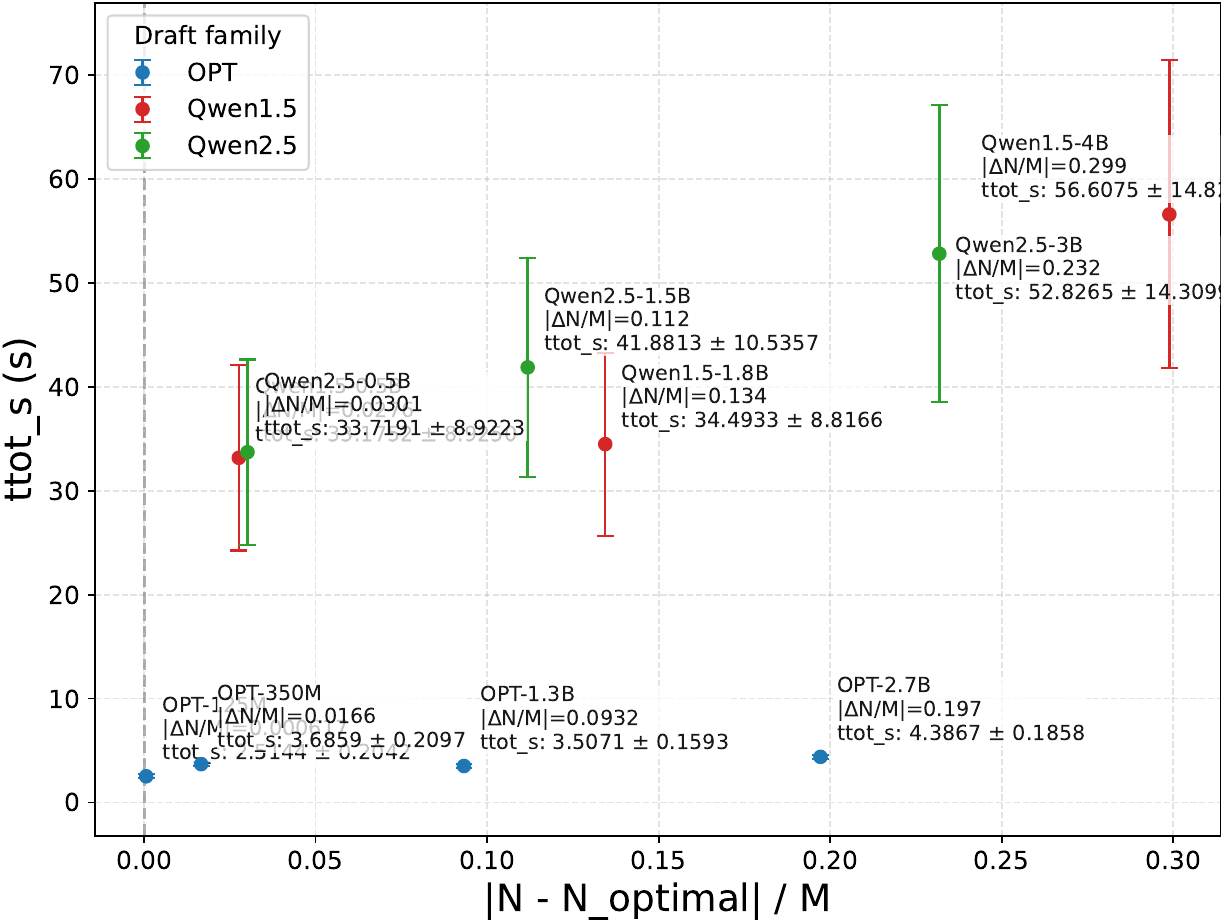}
        \caption{Measured total generation time (TTOT) for generating 250 tokens under speculative decoding for OPT-13B as a function of draft model size \(N\).}
        \label{fig:OPT_13B_TTOT}
    \end{subfigure}
    \\
    \begin{subfigure}[b]{\linewidth}
        \centering
        \includegraphics[width=\linewidth]{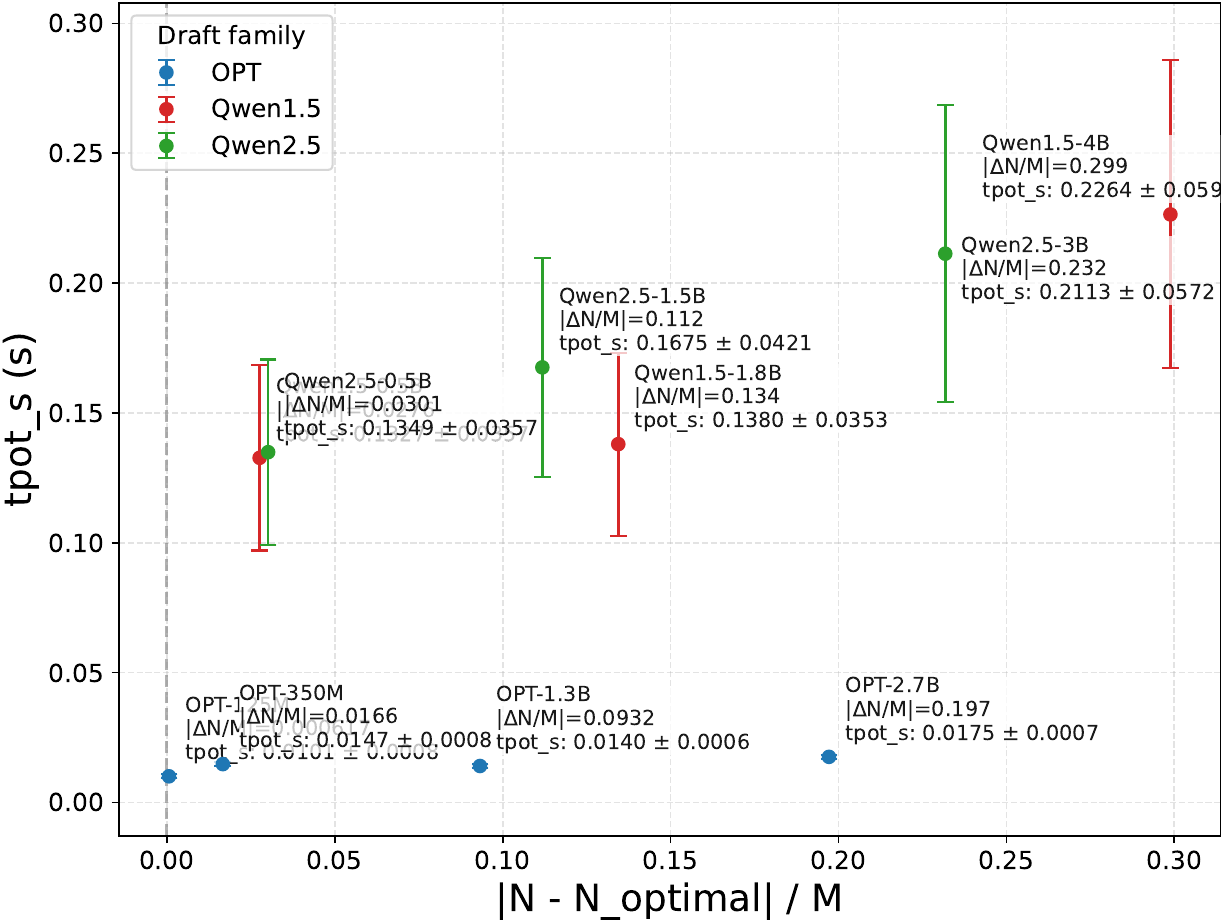}
        \caption{Measured time-per-output-token (TPOT) under speculative decoding for OPT-13B as a function of draft model size \(N\).}
        \label{fig:OPT_13B_TPOT}
    \end{subfigure}
    
    \caption{
    Latency metrics (TTFT, TTOT, and TPOT) for the OPT-13B target model plotted as a function of the normalized deviation \(|N - N^\ast| / M\) from the analytically predicted optimal draft size.
    Each point corresponds to an individual draft model from the OPT, Qwen1.5, or Qwen2.5 families, with error bars indicating 95\% confidence intervals across prompts.The vertical dashed line marks the predicted optimum \(N^\ast\).Across all metrics and draft families, latency increases with distance from \(N^\ast\), supporting the accuracy of the throughput-based optimal draft size prediction.
    }
    \label{fig:Latency-OPT-13B}
\end{figure}

Importantly, Figure~\ref{fig:Latency-OPT-13B} shows that this trend holds consistently across heterogeneous draft families, including OPT, Qwen1.5, and Qwen2.5. When latency is plotted as a function of the normalized deviation \(|N - N^\ast|/M\), draft models from all families exhibit their lowest TTFT, TTOT, and TPOT values at or near \(|N - N^\ast|/M = 0\), with latency increasing as draft size deviates further from the analytically predicted optimum. This alignment indicates that the predicted \(N^\ast\) reliably identifies the vicinity of the latency-minimizing draft size for OPT-13B, despite architectural and tokenizer differences across draft families.

As shown in both Figure~\ref{fig:Latency-OPT-13B} and Table~\ref{table:latency_validation_opt13b}, minor discrepancies between the analytically predicted optimum and the empirically best-performing draft model are observable, particularly for very small or very large draft sizes. These deviations are expected and can be attributed to system-level effects not explicitly captured by Equation~\eqref{eq:throughput_N}, including kernel launch overheads, memory bandwidth saturation, and the fixed execution cost of the draft model. Importantly, these effects influence the magnitude of latency but do not shift the overall location of the latency minimum.



\begin{table*}[h!]
\centering
\begin{tabular}{|c|c|c|c|c|c|}
\hline
\textbf{Parameter}  & \textbf{Estimate}  & \textbf{Std. Error}  & \textbf{Margin of Error}  & \textbf{CI Lower (95\%)}  & \textbf{CI Upper (95\%)} \\
\hline
A & \( -0.0067 \) & \( 0.000607 \) & \( \pm 0.001201 \) & \( -0.007901 \) & \( -0.005499 \) \\
\hline
B & \( 0.012971 \) & \( 0.001545 \) & \( \pm 0.003056 \) & \( 0.009914 \) & \( 0.016027 \) \\
\hline
C & \( 0.642084 \) & \( 0.021228 \) & \( \pm 0.042006 \) & \( 0.600078 \) & \( 0.684090 \) \\
\hline
\multicolumn{6}{|c|}{\textbf{Model Performance}} \\
\hline
\multicolumn{3}{|c|}{\textbf{Mean Squared Error (MSE)}} 
& \multicolumn{3}{c|}{\( 0.001284 \)} \\
\hline
\multicolumn{3}{|c|}{\textbf{\( R^2 \)}} 
& \multicolumn{3}{c|}{\( 0.602296 \)} \\
\hline
\end{tabular}
\caption{Model coefficients for Equation~\ref{eq:plane}, including standard errors, margins of error, and 95\% confidence intervals, along with overall model performance metrics (mean squared error and $R^2$).}
\label{table:fitted_plane}
\end{table*}


\begin{table*}[ht]
\centering
\resizebox{\textwidth}{!}{ 
        \begin{tabular}{|l|l|c|c|c|c|c|c|c|c|}
\hline
\textbf{Target Model} & \textbf{Function} & \textbf{Param 1} & \textbf{MoE 1} & \textbf{CI 1} & \textbf{Param 2} & \textbf{MoE 2} & \textbf{CI 2} & \textbf{MSE} & \textbf{$R^2$} \\
\hline
 & Linear & -0.008262 & 0.001653 & [-0.009915, -0.006609] & 0.832953 & 0.031375 & [0.801578, 0.864329] & 0.000113 & 0.943203 \\
\large{Qwen1.5-14B} & Logarithmic & -0.165206 & 0.026300 & [-0.191506, -0.138905] & 1.155850 & 0.075721 & [1.080129, 1.231571] & 0.000073 & 0.963273 \\
 & Power-law & 1.382782 & 0.158956 & [1.223826, 1.541738] & -0.247168 & 0.040405 & [-0.287573, -0.206763] & 0.000073 & 0.963076 \\
\hline
 & Linear & -0.007686 & 0.001206 & [-0.008892, -0.006480] & 0.804857 & 0.022886 & [0.781971, 0.827743] & 0.000060 & 0.964297 \\
\large{Qwen1.5-32B} & Logarithmic & -0.153441 & 0.017053 & [-0.170494, -0.136388] & 1.104542 & 0.049097 & [1.055444, 1.153639] & 0.000031 & 0.981757 \\
 & Power-law & 1.302398 & 0.100870 & [1.201528, 1.403268] & -0.235315 & 0.027208 & [-0.262523, -0.208108] & 0.000032 & 0.981142 \\
\hline
 & Linear & -0.007030 & 0.001175 & [-0.008205, -0.005854] & 0.795392 & 0.022306 & [0.773086, 0.817698] & 0.000057 & 0.959649 \\
\large{Qwen1.5-72B} & Logarithmic & -0.140489 & 0.016717 & [-0.157205, -0.123772] & 1.069918 & 0.048130 & [1.021789, 1.118048] & 0.000029 & 0.979144 \\
 & Power-law & 1.231728 & 0.091181 & [1.140547, 1.322910] & -0.214429 & 0.025980 & [-0.240409, -0.188449] & 0.000029 & 0.979229 \\
\hline
 & Linear & -0.007472 & 0.001156 & [-0.008627, -0.006316] & 0.735836 & 0.021937 & [0.713899, 0.757774] & 0.000055 & 0.965252 \\
\large{Qwen1.5-110B} & Logarithmic & -0.149112 & 0.016424 & [-0.165536, -0.132688] & 1.027021 & 0.047286 & [0.979735, 1.074306] & 0.000028 & 0.982076 \\
 & Power-law & 1.238509 & 0.103323 & [1.135185, 1.341832] & -0.253840 & 0.029332 & [-0.283173, -0.224508] & 0.000030 & 0.981240 \\
\hline
 & Linear & -0.006945 & 0.001348 & [-0.008293, -0.005597] & 0.798363 & 0.025582 & [0.772782, 0.823945] & 0.000075 & 0.946381 \\
\large{Qwen2.5-14B} & Logarithmic & -0.138591 & 0.022254 & [-0.160845, -0.116337] & 1.068989 & 0.064071 & [1.004917, 1.133060] & 0.000052 & 0.962659 \\
 & Power-law & 1.224068 & 0.120964 & [1.103104, 1.345032] & -0.209879 & 0.034674 & [-0.244554, -0.175205] & 0.000053 & 0.962049 \\
\hline
 & Linear & -0.008567 & 0.001877 & [-0.010444, -0.006690] & 0.817586 & 0.035625 & [0.781960, 0.853211] & 0.000145 & 0.932654 \\
\large{Qwen2.5-32B} & Logarithmic & -0.171822 & 0.029227 & [-0.201049, -0.142594] & 1.153888 & 0.084149 & [1.069739, 1.238037] & 0.000090 & 0.958285 \\
 & Power-law & 1.414605 & 0.183098 & [1.231507, 1.597702] & -0.266090 & 0.045535 & [-0.311625, -0.220556] & 0.000087 & 0.959765 \\
\hline
 & Linear & -0.008433 & 0.001991 & [-0.010425, -0.006442] & 0.808611 & 0.037793 & [0.770817, 0.846404] & 0.000164 & 0.922622 \\
\large{Qwen2.5-72B} & Logarithmic & -0.169348 & 0.031547 & [-0.200895, -0.137801] & 1.140266 & 0.090828 & [1.049438, 1.231094] & 0.000105 & 0.950383 \\
 & Power-law & 1.396431 & 0.196270 & [1.200161, 1.592701] & -0.265035 & 0.049443 & [-0.314478, -0.215592] & 0.000100 & 0.952501 \\
\hline
 & Linear & -0.008005 & 0.001788 & [-0.009793, -0.006218] & 0.831441 & 0.033928 & [0.797514, 0.865369] & 0.000132 & 0.930224 \\
\large{LLaMa3-70B} & Logarithmic & -0.160434 & 0.028584 & [-0.189018, -0.131850] & 1.145350 & 0.082296 & [1.063054, 1.227646] & 0.000086 & 0.954422 \\
 & Power-law & 1.357741 & 0.165924 & [1.191817, 1.523664] & -0.239126 & 0.042938 & [-0.282064, -0.196187] & 0.000084 & 0.955649 \\
\hline
 & Linear & -0.006045 & 0.001108 & [-0.007153, -0.004937] & 0.779966 & 0.021028 & [0.758937, 0.800994] & 0.000051 & 0.951894 \\
\large{LLaMa3.1-70B} & Logarithmic & -0.120517 & 0.018271 & [-0.138788, -0.102246] & 1.015201 & 0.052604 & [0.962596, 1.067805] & 0.000035 & 0.966577 \\
 & Power-law & 1.129190 & 0.091637 & [1.037553, 1.220828] & -0.182634 & 0.028439 & [-0.211073, -0.154196] & 0.000036 & 0.965962 \\
\hline
 & Linear & -0.004219 & 0.001611 & [-0.005830, -0.002608] & 0.732211 & 0.030576 & [0.701635, 0.762787] & 0.000107 & 0.820131 \\
\large{OPT-13B} & Logarithmic & -0.079686 & 0.037777 & [-0.117463, -0.041910] & 0.883694 & 0.108763 & [0.774931, 0.992457] & 0.000150 & 0.747328 \\
 & Power-law & 0.926916 & 0.159383 & [0.767533, 1.086300] & -0.121164 & 0.060080 & [-0.181243, -0.061084] & 0.000157 & 0.737010 \\
\hline
 & Linear & -0.004025 & 0.000950 & [-0.004975, -0.003075] & 0.801596 & 0.018032 & [0.783564, 0.819628] & 0.000037 & 0.922662 \\
\large{OPT-30B} & Logarithmic & -0.077500 & 0.024004 & [-0.101504, -0.053496] & 0.950355 & 0.069110 & [0.881244, 1.019465] & 0.000061 & 0.873879 \\
 & Power-law & 0.987912 & 0.097590 & [0.890322, 1.085502] & -0.106544 & 0.034491 & [-0.141035, -0.072053] & 0.000064 & 0.866974 \\
\hline
 & Linear & -0.004553 & 0.000804 & [-0.005357, -0.003749] & 0.850306 & 0.015264 & [0.835042, 0.865571] & 0.000027 & 0.955174 \\
\large{OPT-66B} & Logarithmic & -0.088115 & 0.022067 & [-0.110182, -0.066048] & 1.019858 & 0.063533 & [0.956325, 1.083391] & 0.000051 & 0.913782 \\
 & Power-law & 1.066712 & 0.093086 & [0.973626, 1.159797] & -0.115104 & 0.030481 & [-0.145586, -0.084623] & 0.000055 & 0.906985 \\
\hline
 & Linear & -0.005858 & 0.001169 & [-0.007027, -0.004689] & 0.704290 & 0.022195 & [0.682095, 0.726486] & 0.000056 & 0.943444 \\
\large{Seed-OSS-36B} & Logarithmic & -0.117008 & 0.019092 & [-0.136100, -0.097916] & 0.932871 & 0.054969 & [0.877903, 0.987840] & 0.000038 & 0.961488 \\
 & Power-law & 1.056024 & 0.099654 & [0.956369, 1.155678] & -0.199176 & 0.033095 & [-0.232271, -0.166081] & 0.000038 & 0.961522 \\
\hline
\end{tabular}
    }
\caption{Complete regression results for linear, logarithmic, and power-law fits of $\alpha$ as a function of draft model perplexity, including parameter estimates, margins of error (MoE), 95\% confidence intervals (CI), mean squared error (MSE), and coefficient of determination ($R^2$).}
\label{table:alpha_regression_draft}
\end{table*}


\begin{table*}[ht]
 
\renewcommand{\arraystretch}{1.1}
    \resizebox{\textwidth}{!}{ 
        \begin{tabular}{|l|l|c|c|c|c|c|}
\hline
\textbf{Target Model} & \textbf{M (target size)} & \textbf{\(D^\prime\) (training target size)}  & \textbf{Draft Family} & \textbf{\(D\) (training draft size)} &  \textbf{ \(N^\ast\)(Optimal N)} & \textbf{\(\mathcal{T}(N^\ast)\)} \\
\hline
 & & & OPT & 180B & 124039820.3 & 8.947e-11  \\
 \large{Qwen1.5-14B} & 14167290880 &  \large{3T} & Qwen1.5 & 2.4T & 115390938.2  & 9.165e-11 \\
 & & & Qwen2.5 & 18T & 112685392.4 & 9.235e-11  \\
\hline
 & & & OPT & 180B & 198605884.7 & 4.061e-11 \\
\large{Qwen1.5-32B}& 32512218112 &  \large{3T} & Qwen1.5 & 2.4T &  184494249.2 & 4.154e-11    \\
 & & & Qwen2.5 & 18T & 180081907.1 &  4.183e-11  \\
\hline
 & & & OPT & 180B & 317168451.4 & 1.883e-11  \\
\large{Qwen1.5-72B} & 72287920128 &  \large{3T}& Qwen1.5 & 2.4T  & 294269636.4 & 1.924e-11 \\
 & & & Qwen2.5 & 18T & 287112608.1 & 1.936e-11   \\
\hline
 & & & OPT & 180B & 410570760.1 & 1.241e-11  \\
\large{Qwen1.5-110B} & 111209914368 & \large{3T} &  Qwen1.5 & 2.4T  & 380698767.3 & 1.267e-11  \\
 & & & Qwen2.5 & 18T & 371364065.7 & 1.275e-11   \\
\hline
 & & & OPT& 180B & 126997571.6 & 8.556e-11 \\
\large{Qwen2.5-14B} & 14770033664 & 18T  & Qwen1.5 & 2.4T  & 118129957.5 & 8.762e-11  \\
 & &  & Qwen2.5 & 18T & 115356193.9 & 8.829e-11   \\
\hline
 & & & OPT & 180B & 199582637.5 & 4.005e-11 \\
\large{Qwen2.5-32B} & 32763876352 & 18T  & Qwen1.5 & 2.4T  & 185393085.4 & 4.095e-11 \\
 & & & Qwen2.5 & 18T & 180956639.7 &  4.124e-11 \\
\hline
 & & & OPT & 180B & 318409631.5 & 1.861e-11  \\
\large{Qwen2.5-72B} & 72706203648 & 18T  & Qwen1.5 & 2.4T & 295409960.3 &  1.900e-11  \\
 & & & Qwen2.5 & 18T & 288221790.3 &  1.912e-11  \\
\hline
 & & & OPT & 180B & 312760391.8 & 1.916e-11 \\
\large{LLaMa3-70B} & 70553706496 & 15T & Qwen1.5 & 2.4T & 290182058.1 & 1.957e-11 \\
 & & & Qwen2.5 & 1.4T & 283125440.3 & 1.970e-11   \\
\hline
 & & & OPT & 180B & 312760391.8 & 1.916e-11  \\
\large{LLaMa3.1-70B} & 70553706496 & 15T &  Qwen1.5 & 2.4T & 290182058.1 & 1.957e-11  \\
 & & & Qwen2.5 & 18T & 283125440.3 & 1.970e-11  \\
\hline
 & & & OPT & 180B & 117313808.6 & 1.009e-10 \\
\large{OPT-13B} & 12853473280 & 180B & Qwen1.5 & 2.4T & 109171489.6 &  1.034e-10 \\
 & & & Qwen2.5 & 18T & 106623620.8 &  1.043e-10 \\
\hline
 & & & OPT & 180B & 189210723.1 & 4.510e-11 \\
\large{OPT-30B} & 29974540288 & 180B & Qwen1.5 & 2.4T & 175814080.3 & 4.616e-11 \\
 & & & Qwen2.5 & 18T & 171624238.6 & 4.650e-11  \\
\hline
 & & & OPT & 180B & 299138626.9 & 2.122e-11   \\
\large{OPT-66B} & 65693122560 & 180B & Qwen1.5 & 2.4T & 277613865.6 & 2.169e-11   \\
 & & & Qwen2.5 & 18T & 270884739.1 & 2.184e-11   \\
\hline
 & & & OPT & 180B & 211245817.4 & 3.649e-11  \\
\large{Seed-OSS-36B} & 36151104512 & 12T & Qwen1.5 & 2.4T & 196196967 & 3.730e-11  \\
 & & & Qwen2.5 & 18T & 191492038.2 & 3.757e-11 \\
\hline
\end{tabular}
    }
       \caption{Let optimal draft model characteristics for each target model. For each target, we report the model size \(M\), training dataset size \(D\), and the optimal draft model size \(N\) that maximizes decoding throughput. The last column reports the achieved throughput in tokens per FLOP for each target model in the measured optimal draft model. This table provides the empirical foundation for our scaling law analysis, illustrating how optimal draft configurations vary with model scale and data.}

    \label{table:Optimal_draft_model_size_throughput}

\end{table*}


\begin{table*}[ht]
\centering
\renewcommand{\arraystretch}{1.15}
\resizebox{\textwidth}{!}{
\begin{tabular}{|l|l|c|c|c|c|c|}
\hline
\textbf{Draft Model} &
\textbf{Draft Family} &
\textbf{\(N^\ast\)(Optimal N)} &
\textbf{\(|N-N^\ast|/M\)} &
\textbf{TTFT (s)} &
\textbf{TTOT (s)} &
\textbf{TPOT (s)} \\
\hline

OPT-125M   & OPT     & 117M & 0.000617 &
$0.04864 \pm 0.0017$ &
$2.514 \pm 0.2042$ &
$0.01006 \pm 0.00082$ \\
\hline

OPT-350M   & OPT     & 117M & 0.01664 &
$0.050003 \pm 0.00126$ &
$3.686 \pm 0.20972$ &
$0.01474 \pm 0.00084$ \\
\hline

OPT-1.3B   & OPT     & 117M & 0.09324 &
$0.04779 \pm 0.00118$ &
$3.507 \pm 0.1593$ &
$0.01403 \pm 0.00064$ \\
\hline

OPT-2.7B   & OPT     & 117M & 0.1972 &
$0.04762 \pm 0.00101$ &
$4.387 \pm 0.1858$ &
$0.01755 \pm 0.00074$ \\
\hline

Qwen1.5-0.5B & Qwen1.5 & 109M & 0.0276 &
$0.1001 \pm 0.00816$ &
$33.18 \pm 8.925$ &
$0.1327 \pm 0.0357$ \\
\hline

Qwen1.5-1.8B & Qwen1.5 & 109M & 0.1344 &
$0.1008 \pm 0.01120$ &
$34.49 \pm 8.817$ &
$0.138 \pm 0.03527$ \\
\hline

Qwen1.5-4B & Qwen1.5 & 109M & 0.2988 &
$0.1212 \pm 0.01622$ &
$56.61 \pm 14.82$ &
$0.2264 \pm 0.05929$ \\
\hline

Qwen2.5-0.5B & Qwen2.5 & 107M & 0.03014 &
$0.1094 \pm 0.00922$ &
$33.72 \pm 8.922$ &
$0.1349 \pm 0.03569$ \\
\hline

Qwen2.5-1.5B & Qwen2.5 & 107M & 0.1118 &
$0.1229 \pm 0.01112$ &
$41.88 \pm 10.54$ &
$0.1675 \pm 0.04214$ \\
\hline

Qwen2.5-3B & Qwen2.5 & 107M & 0.2318 &
$0.1201 \pm 0.01477$ &
$52.83 \pm 14.31$ &
$0.2113 \pm 0.05724$ \\
\hline

\end{tabular}
}
\caption{
Latency-based validation of the analytically predicted optimal draft size \(N^\ast\) for speculative decoding with an OPT-13B target model.
For each evaluated draft model, we report the predicted optimum \(N^\ast\), the normalized distance \(|N-N^\ast|/M\), and empirical latency measurements obtained from speculative decoding on 50 HellaSwag prompts.
Latency metrics include time-to-first-token (TTFT), total generation time (TTOT), and time per output token (TPOT), reported as mean $\pm$ margin of error corresponding to 95\% confidence intervals.
}
\label{table:latency_validation_opt13b}
\end{table*}

\begin{table*}[ht]

\centering
\resizebox{0.85\textwidth}{!}{ 
\begin{tabular}{|c|c|c|c|}
\hline
\textbf{Parameter} & \textbf{Range} & \textbf{Number of Grid Points} & \textbf{Description} \\ \hline
\( N \) & \( 10^8 \) to \(10^{10}\) & 10000&Initial draft model size  \\ \hline
\( M \) & \(  13 \times 10^{9} \) to \(  110 \times 10^{9} \) & 8 &Target model size  \\ \hline
\( D \) & \( 10^{12} \) to \( 10^{13} \) & 6 &Number of tokens used in draft model training \\ \hline
\( D\prime \) & \( 10^{12} \) to \( 10^{13} \) & 6 & Number of training tokens used in target model training \\ \hline

\end{tabular}
}
\caption{Ranges for \( N \), \( D \), and \( M \) used in the analysis.}
\label{table:ranges}

\end{table*}

\begin{table*}[ht]

\centering
\resizebox{0.95\textwidth}{!}{ 
\begin{tabular}{|c|c|c|c|c|}
\hline
\textbf{Parameter} 
& \textbf{Estimate} 
& \textbf{MoE (95\%)} 
& \textbf{95\% CI (lower)} 
& \textbf{95\% CI (upper)} \\ \hline

$\mu$ 
& $4.89\times 10^{-3}$ 
& $7.64\times 10^{-4}$ 
& $4.12\times 10^{-3}$ 
& $5.65\times 10^{-3}$ \\ \hline

$M_0$ 
& $7.23\times 10^{7}$ 
& $1.46\times 10^{6}$ 
& $7.09\times 10^{7}$ 
& $7.38\times 10^{7}$ \\ \hline

$\gamma$ 
& $-6.07\times 10^{-5}$ 
& $1.79\times 10^{-5}$ 
& $-7.86\times 10^{-5}$ 
& $-4.28\times 10^{-5}$ \\ \hline

$\gamma'$ 
& $1.16\times 10^{-6}$ 
& $1.78\times 10^{-5}$ 
& $-1.67\times 10^{-5}$ 
& $1.90\times 10^{-5}$ \\ \hline

\hline
$R^2$ 
& \multicolumn{4}{c|}{$0.978$} \\ \hline

Adj. $R^2$ 
& \multicolumn{4}{c|}{$0.978$} \\ \hline

Draft-size search grid 
& \multicolumn{4}{c|}{$|N| = 10000$ (log-spaced; evaluated only for $N < 10B$)} \\ \hline
Number of Grid Points 
& \multicolumn{4}{c|}{$|M| = 8,\; |D| = 6,\; |D'| = 6$} \\ \hline

Observations 
& \multicolumn{4}{c|}{$288$} \\ \hline

Regression type 
& \multicolumn{4}{c|}{OLS with HC3 robust standard errors} \\ \hline

\end{tabular}
}
\caption{Scaling-law regression results for the throughput-optimal draft size.
The model
$N^\*/M = \mu + M_0/M + \gamma \log D + \gamma' \log D'$
is fit using ordinary least squares with heteroskedasticity-robust (HC3) standard errors.
Margins of error (MoE) correspond to half-widths of the reported 95\% confidence intervals.}
\label{table:nstar_scaling_moe_first}

\end{table*}

\begin{table*}[ht]

\centering
\resizebox{0.85\textwidth}{!}{ 
\begin{tabular}{|c|c|c|c|c|}
\hline
\textbf{Parameter} 
& \textbf{Estimate} 
& \textbf{MoE (95\%)} 
& \textbf{95\% CI (lower)} 
& \textbf{95\% CI (upper)} \\ \hline

$\mu$ 
& $2.71\times 10^{-3}$ 
& $4.38\times 10^{-5}$ 
& $2.67\times 10^{-3}$ 
& $2.75\times 10^{-3}$ \\ \hline

$M_0$ 
& $8.71\times 10^{7}$ 
& $2.18\times 10^{6}$ 
& $8.50\times 10^{7}$ 
& $8.93\times 10^{7}$ \\ \hline

\hline
$R^2$ 
& \multicolumn{4}{c|}{$0.9865$} \\ \hline

Observations 
& \multicolumn{4}{c|}{$288$} \\ \hline

Number of Grid Points 
& \multicolumn{4}{c|}{$|M|=8,\;|D|=6,\;|D'|=6$} \\ \hline

Regression type 
& \multicolumn{4}{c|}{OLS with HC3 robust standard errors} \\ \hline

\end{tabular}
}
\caption{Pooled regression results for the leading-order scaling law
$N^\* = \mu M + M_0$.
The regression is performed over all $(M,D,D')$ configurations, treating
dataset-induced variation as part of the residual.
Margins of error (MoE) correspond to half-widths of the reported 95\% confidence intervals.}
\label{table:nstar_linear_pooled}

\end{table*}


\begin{figure*}[t]
    \centering

    \begin{subfigure}[t]{0.48\linewidth}
        \centering
        \includegraphics[width=\linewidth,
  height=0.255\textheight,
  keepaspectratio]{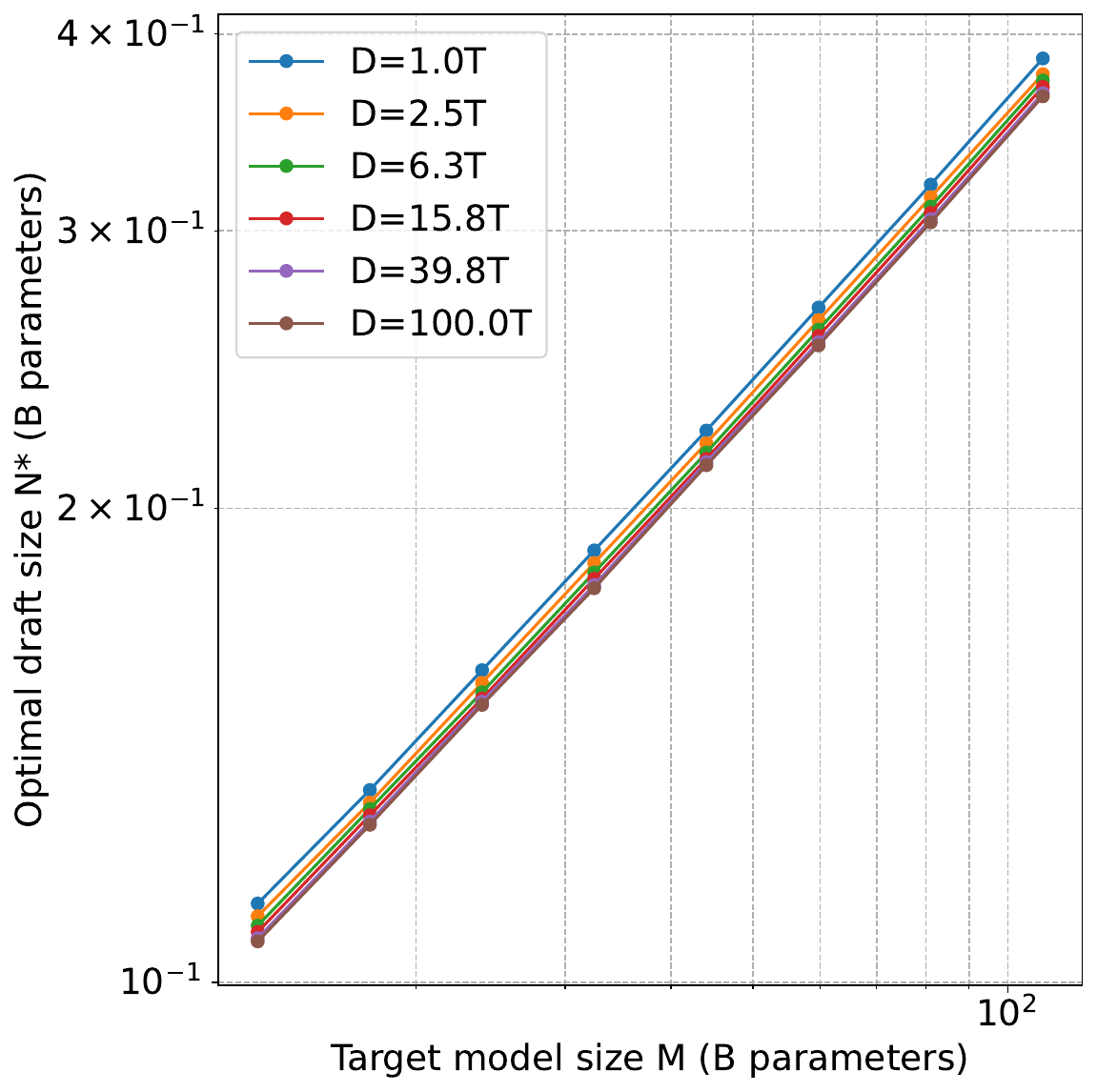}
        \caption{
        Optimal draft size \(N^\ast\) as a function of target model size \(M\),
        shown for multiple draft training dataset sizes \(D\) while fixing
        the target training dataset size \(D^\prime\).
        }
        \label{fig:Nstar_vs_M_raw}
    \end{subfigure}
    \hfill
    \begin{subfigure}[t]{0.48\linewidth}
        \centering
        \includegraphics[width=\linewidth,
  height=0.255\textheight,
  keepaspectratio]{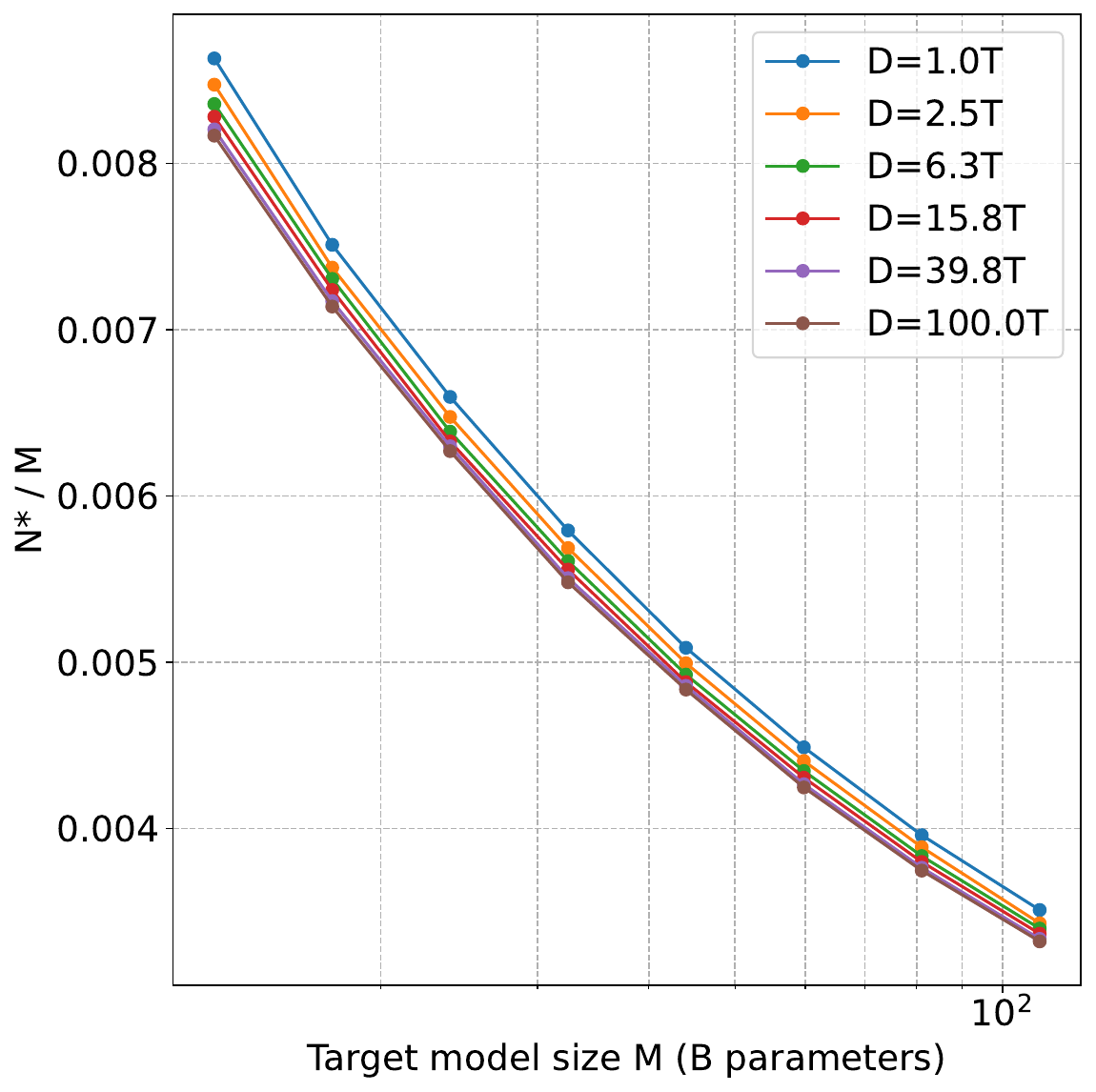}
        \caption{
        Normalized optimal draft size \(N^\ast/M\) as a function of \(M\),
        corresponding to Fig.~\ref{fig:Nstar_vs_M_raw}.
        }
        \label{fig:Nstar_over_M_vs_M}
    \end{subfigure}


    \begin{subfigure}[t]{0.48\linewidth}
        \centering
        \includegraphics[width=\linewidth,
  height=0.255\textheight,
  keepaspectratio]{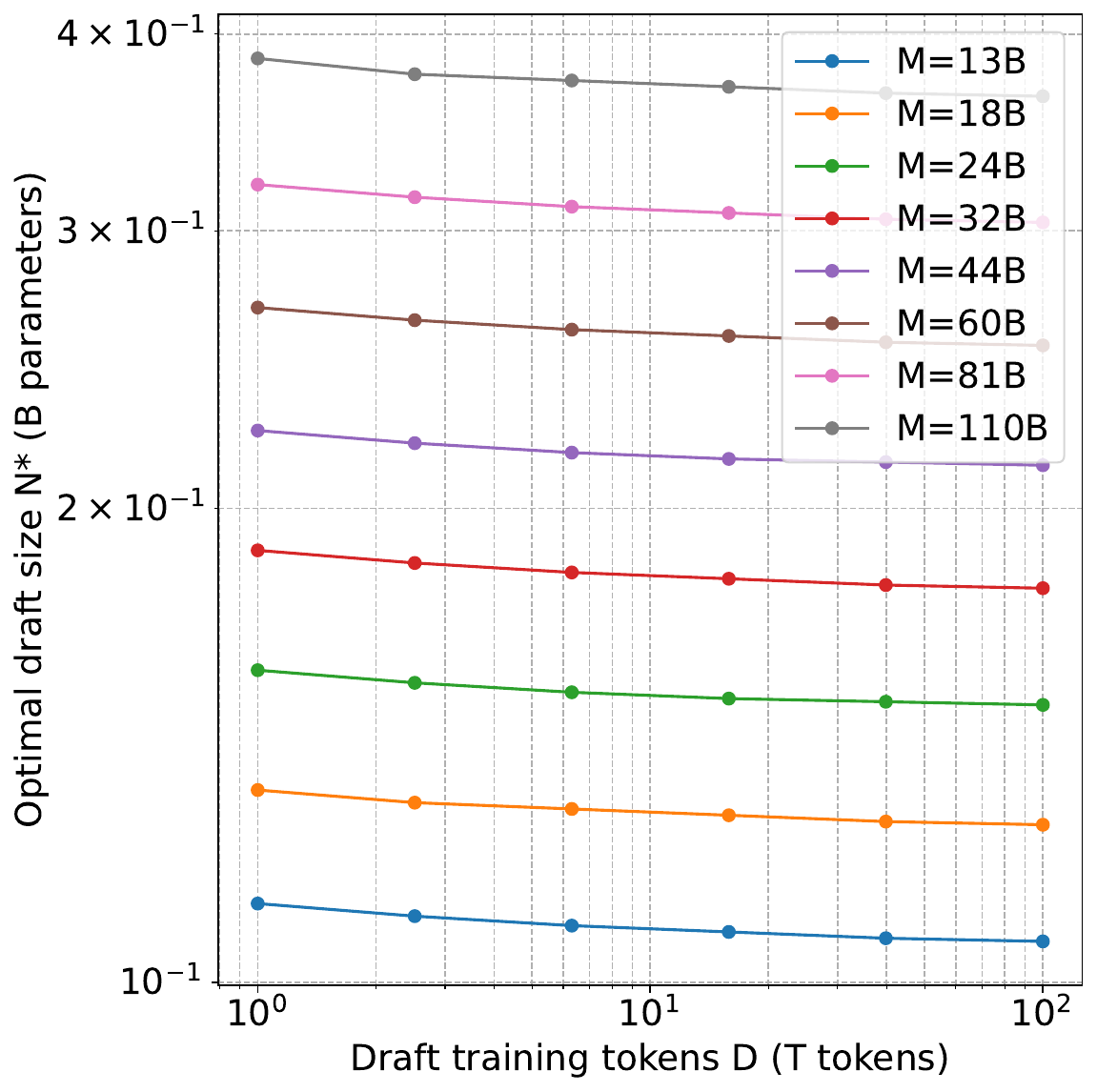}
        \caption{
        Optimal draft size \(N^\ast\) as a function of draft training dataset
        size \(D\), shown for multiple target model sizes \(M\) while fixing
        \(D^\prime\).
        }
        \label{fig:Nstar_vs_D_raw}
    \end{subfigure}
    \hfill
    \begin{subfigure}[t]{0.48\linewidth}
        \centering
        \includegraphics[width=\linewidth,
  height=0.255\textheight,
  keepaspectratio]{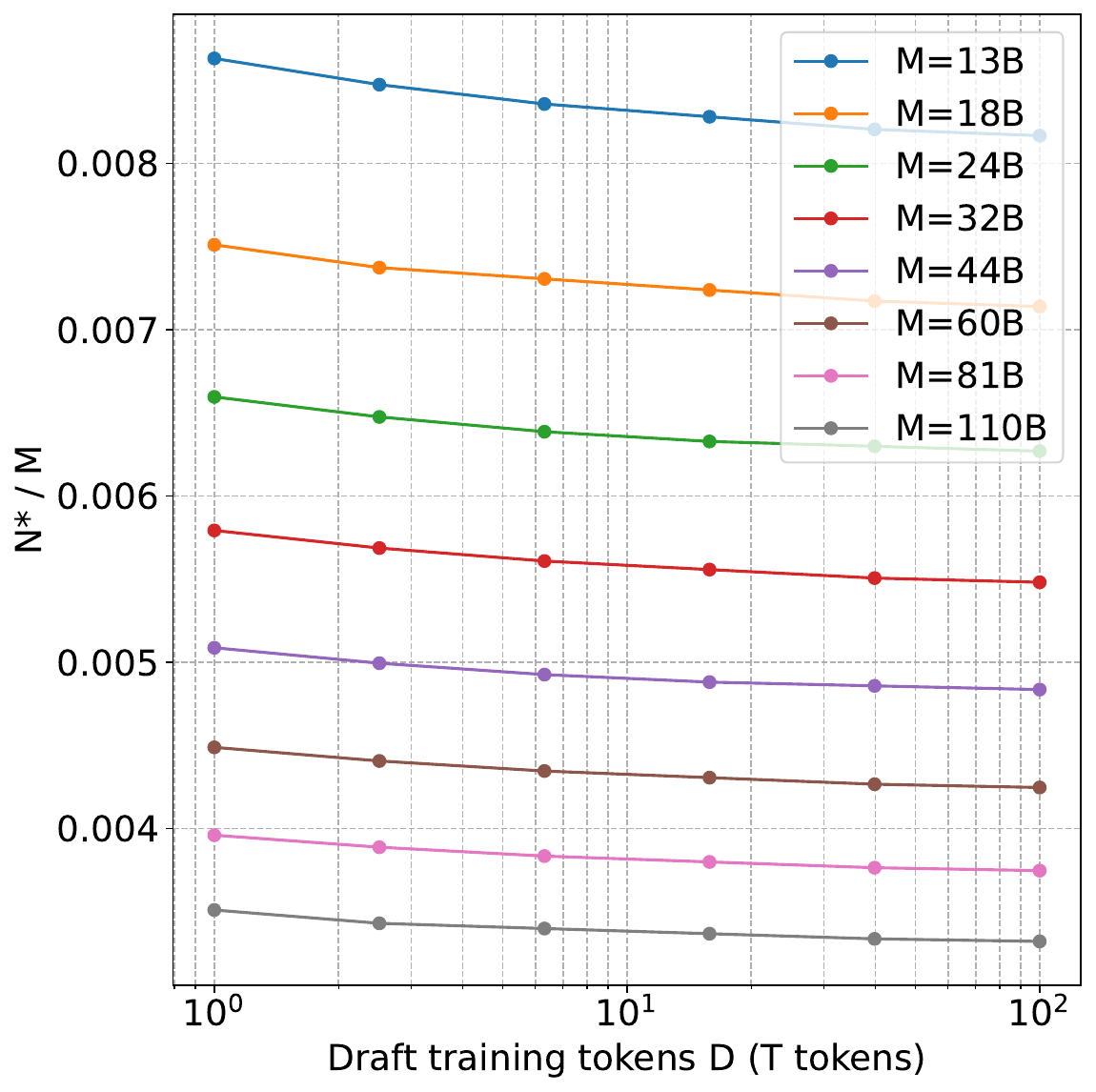}
        \caption{
        Normalized optimal draft size \(N^\ast/M\) as a function of \(D\),
        corresponding to Fig.~\ref{fig:Nstar_vs_D_raw}.
        }
        \label{fig:Nstar_over_M_vs_D}
    \end{subfigure}


    \begin{subfigure}[t]{0.48\linewidth}
        \centering
        \includegraphics[width=\linewidth,
  height=0.255\textheight,
  keepaspectratio]{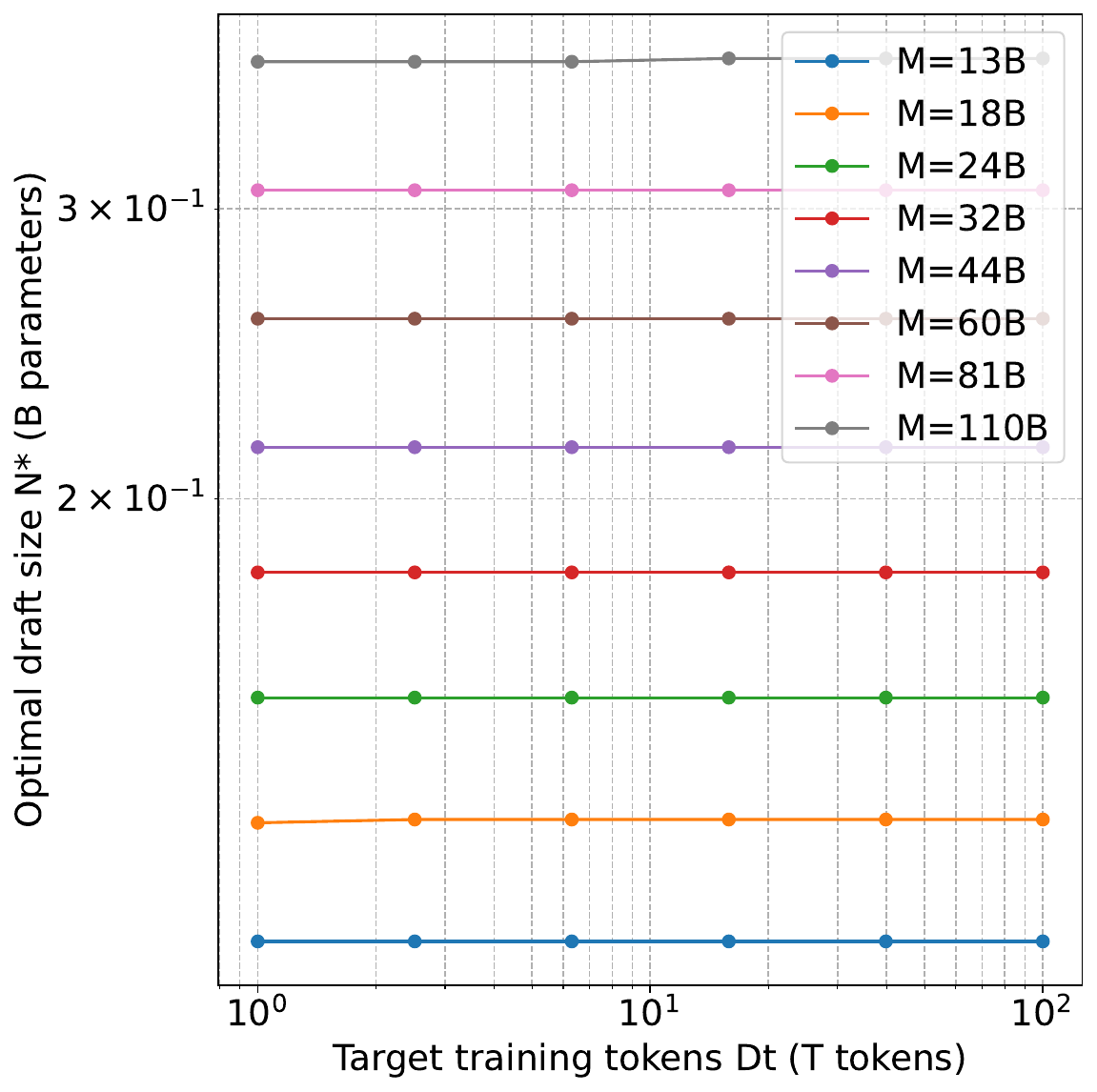}
        \caption{
        Optimal draft size \(N^\ast\) as a function of the target training
        dataset size \(D^\prime\), shown for multiple target model sizes \(M\)
        while fixing the draft training dataset size \(D\).
        }
        \label{fig:Nstar_vs_Dt_raw}
    \end{subfigure}
    \hfill
    \begin{subfigure}[t]{0.48\linewidth}
        \centering
        \includegraphics[width=\linewidth,
  height=0.255\textheight,
  keepaspectratio]{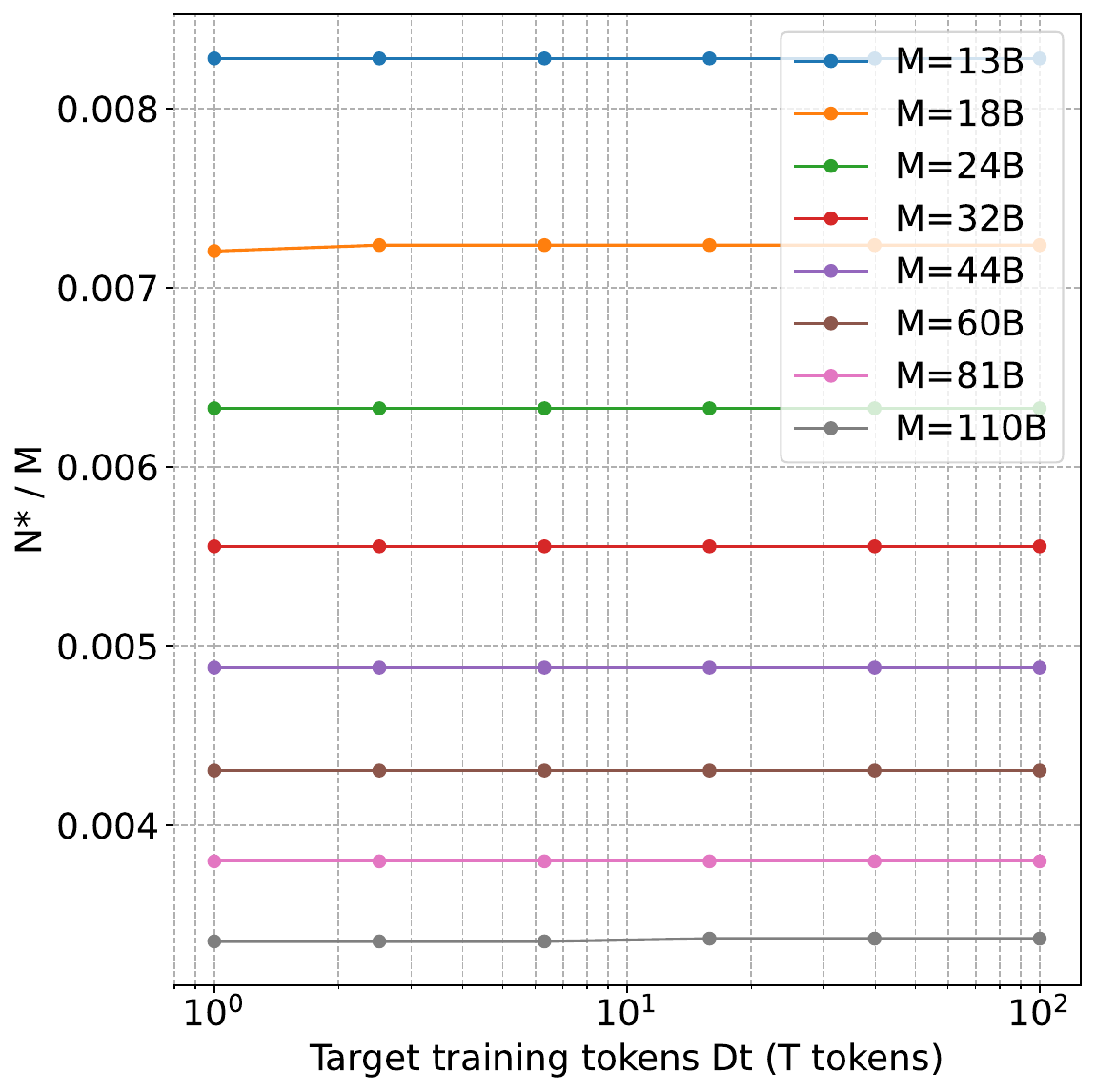}
        \caption{
        Normalized optimal draft size \(N^\ast/M\) as a function of \(D^\prime\),
        corresponding to Fig.~\ref{fig:Nstar_vs_Dt_raw}.
        }
        \label{fig:Nstar_over_M_vs_Dt}
    \end{subfigure}

    \caption{
    Numerical characterization of the throughput-optimal draft model size
    \(N^\ast\) as a function of the target model size \(M\), the draft training
    dataset size \(D\), and the target training dataset size \(D^\prime\).
    Each row presents the raw dependence of \(N^\ast\) (left) alongside the
    corresponding normalized view \(N^\ast/M\) (right).
    The results show that the dominant scaling of the optimal draft size is
    approximately linear in the target model size, while the effects of the
    training dataset sizes act as weaker, second-order corrections.
    }
    \label{fig:Nstar_scaling_summary}
\end{figure*}


\end{document}